\documentclass[journal]{IEEEtran}
\usepackage{amsmath,amssymb,bm}
\usepackage{graphicx}
\usepackage{tikz}
\usetikzlibrary{arrows.meta,positioning,calc,shapes.geometric}
\usepackage{booktabs}
\usepackage{cite}
\usepackage{url}

\usepackage{xcolor}
\usepackage{colortbl}

\usepackage{multirow}

\usepackage{subcaption}

\newcommand{\twoval}[2]{\begin{tabular}[c]{@{}c@{}}#1\\#2\end{tabular}}

\title{Reference-Twist-Based Velocity-Domain Spatiotemporal Calibration \\ for mmWave Radar, Event Camera, and IMU}

\author{Author Names Omitted for Review}
\author{Zhenghao Zou, Yang~Lyu*,~\IEEEmembership{Member,~IEEE}, Xiaohu Guo, Chunhui Zhao, Quan Pan,~\IEEEmembership{Member,~IEEE}

}

\begin{document}
\maketitle

\begin{abstract}
Multi-sensor navigation requires a common spatial and temporal basis, but geometric calibration relies on targets, features, point clouds, or pose associations that are difficult to establish across mmWave radars, event cameras, and inertial measurement units (IMUs). These sensors nevertheless share rigid-body motion. This paper formulates reference-conditioned velocity-domain spatiotemporal calibration for a rigid suite of two mmWave radars, one event camera, and two IMUs. Given a trusted reference body twist, radar Doppler, event normal flow, and inertial measurements are regressed against the same motion to estimate relative rotations, lever arms, time offsets, and sensor-specific bias and scale without artificial targets, structure from motion, or cross-modal point-cloud registration. The reference is treated as a conditioning input rather than a sensor to be calibrated. We compare joint and reference-conditioned estimation, derive conditional identifiability and CRLB-level covariance and bias bounds, and state the operating conditions under which conditioning is informative. Indoor and forest recordings with axis-separated motion evaluate initialization, temporal alignment, motion degeneracy, reference-input sensitivity, and residual consistency. Because the reference does not add physical observability, the estimates remain conditional on reference accuracy, synchronization, motion excitation, and projection diversity.
\end{abstract}

\section{Introduction}
Robot navigation relies on multi-source state feedback for localization, planning, and control. Multi-sensor fusion is a standard way to obtain such feedback with complementary robustness, but fusion is meaningful only after the measurements have been expressed in a common spatial and temporal basis. Spatial calibration determines relative orientations and lever arms; temporal calibration associates each sample with the same body motion. Without this basis, residuals from different sensors describe different states at different instants, and their apparent complementarity cannot be exploited consistently \cite{river22025,li2024lirio,dong2025ekfrio}.

Velocity provides a compact state feedback for establishing this basis. Unlike a global pose, velocity is a local quantity: it does not require a map or global position, and high-rate linear and angular velocity constraints remain informative in motion estimation and in directions where geometric observability degenerates. For heterogeneous sensing, velocity-domain observations are also easier to relate than geometric features, because rigid-body kinematics provides a common physical model even when different sensors observe disjoint scene content. The radar, event camera, and IMU considered here are all velocity-sensitive: Doppler measures radial velocity, event normal flow encodes image-plane motion, and inertial measurements provide angular rate and specific force. None alone supplies a complete body twist, but their native observations share motion as a linking variable.

Geometry-based calibration is widely used when targets, features, point clouds, or pose graphs can be associated across sensors \cite{furgale2013unified,he2026radarcam,li2024multilidar}. These associations are fragile for the present suite. Radar returns are sparse and susceptible to multipath; event streams report asynchronous brightness changes rather than persistent appearance \cite{gallego2022event}; and an IMU provides no geometric observation of the scene. Consequently, a radar detection, an event, and an inertial sample have no natural common geometric correspondence, even when each sensor is individually well modeled. Enforcing geometry-based calibration would require additional targets or scene structure and a reliable cross-modal association stage, which the heterogeneous measurements do not by themselves provide.

Motion-based calibration offers an alternative when direct cross-modal correspondence is unavailable \cite{qiu2020motioncorr,chen2023multicam}. Rigidly mounted sensors share the same body motion, so motion correlation can recover rotation and timing without matching scene features, while continuous-time formulations can represent motion between asynchronous samples. Trajectory alignment is one route, but it requires poses or integrated trajectories and therefore inherits drift, gravity, and initialization errors. Velocity-domain calibration instead compares native velocity observations directly: it avoids pose integration and global attitude propagation while retaining the lever-arm term that couples rotation to translation. The proposed formulation follows this second route.

We further adopt a reference-conditioned setting: given a trusted reference body twist, each sensor is regressed against the same body motion, converting the calibration problem into modality-specific parameter blocks. When the reference is independent, synchronized, and sufficiently excited, this offers an effective engineering route to rapid and accurate spatiotemporal calibration: the reference supplies the common exogenous motion that would otherwise be estimated as a shared latent state, breaks the information isolation between heterogeneous sensor blocks, and avoids propagating the degeneracy and errors of a jointly estimated velocity state into every sensor-specific block. The reference is an input rather than a sensor to be calibrated; it changes the estimation structure but does not add physical observability. Its operating conditions and CRLB-level gain are analyzed in Section~\ref{sec:reference_decoupling}.

\begin{figure}[!t]
    \centering
    \includegraphics[width=\columnwidth]{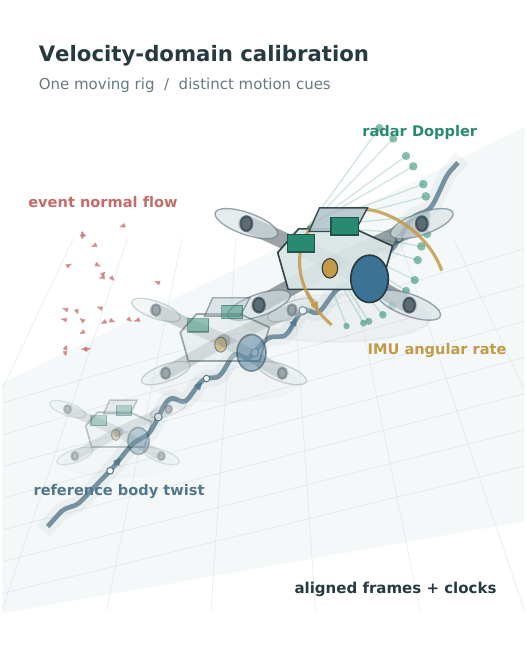}
    \caption{Motion-centric view of reference-conditioned spatiotemporal calibration. A moving multisensor rig links radar Doppler, event normal flow, and IMU angular-rate observations to a common reference body twist, aligning their frames and clocks. Conceptual rendering: the reference-speed profile and event-flow directions use Sequence~3 data; the sensor silhouette and radial Doppler samples illustrate the measurement geometry rather than quantitative results.}
    \label{fig:figure1}
\end{figure}

The proposed formulation connects each sensor's native observation to the same reference twist, retains the lever-arm contribution under rotational excitation, and preserves modality-specific measurement models. The resulting estimates remain conditional on reference quality, synchronization, and motion excitation.

The contributions of this work are summarized as follows:
\begin{itemize}
\item We analyze the estimation structure, identifiability conditions, and CRLB-level gain boundary of reference-conditioned velocity-domain spatiotemporal calibration, clarifying when a trusted reference removes shared-state uncertainty and when native measurement degeneracy remains.
\item We propose a reference-twist-based radar--event--IMU spatiotemporal calibration framework that integrates temporal alignment, rotation alignment, lever-arm estimation, and joint refinement through modality-specific residuals.
\item We develop a body-frame gravity-decoupled inertial constraint that removes explicit world-gravity and attitude integration from the accelerometer residual, reducing sensitivity to global attitude propagation.
\item We collect and evaluate indoor and forest recordings with axis-separated omnidirectional excitation, and report initialization, timing, motion-degeneracy, reference-input sensitivity, and residual-consistency results on a rigid radar--event--IMU platform.
\end{itemize}

The scope is deliberately defined by the availability of a trusted reference, not by a claim that one calibration architecture dominates all others. The proposed method is not a replacement for reference-free joint calibration; joint estimators remain the appropriate choice when no reliable reference is available. Rather, it is a practical and theoretically grounded choice for reference-available settings, where it offers an effective engineering route to fast, accurate spatiotemporal calibration without depending on a jointly estimated velocity state. The reference is a conditioning input that simplifies the estimator structure; it does not add physical observability. The analysis and experiments therefore characterize the operating envelope of this choice rather than establish a universal ranking of calibration methods.

\section{Related Work}
\subsection{Geometry-Based Spatiotemporal Calibration}
Geometry-based calibration estimates sensor relationships from identifiable targets, natural scene structure, or temporal frame-to-frame associations. Its common requirement is an association that can be expressed geometrically across sensors or over time. Furgale et al. jointly estimate camera--IMU extrinsics and timing using a static calibration pattern and continuous-time batch estimation \cite{furgale2013unified}. Lee et al. use target trajectories to align radar and LiDAR, with different target and map requirements in factory and on-road settings \cite{lee2023radarlidar}. These methods provide geometric constraints when the relevant observations can be identified and associated across the participating sensors.

MultiCal extends target-based calibration to multiple cameras, LiDARs, and IMUs using several planar targets, including configurations with non-overlapping fields of view \cite{zhi2022multical}. Such designs show that a shared target need not be observed by every sensor, although target geometry and observation models remain part of the calibration. Targetless methods instead exploit natural scene structure. Li et al. estimate the parameters of a heterogeneous camera--IMU--LiDAR system using continuous-time batch estimation \cite{li2023hetero}. For radar--camera calibration, SPECal relates separately estimated sensor poses and uses camera depth for refinement \cite{specal2024}. Ge et al. combine moving-object contours with radar Doppler and camera optical flow \cite{ge2023automatic}, illustrating that geometric and motion constraints can also be complementary.

Recent methods adapt geometric constraints to the measurements available in each modality. Radar calibration using corner reflectors and multiple platform poses jointly estimates spatial alignment and timing \cite{jung2026spatiotemporal}, while uncertainty-aware radar--camera calibration models the propagation of radar measurement errors into geometric registration \cite{cao20254d}. L2E~\cite{ta2023l2e} directly aligns LiDAR returns with event-camera observations for spatial calibration. Such results demonstrate that heterogeneous observations can support geometric calibration when suitable associations or scene constraints are available; they do not imply that the same constraints transfer directly to sparse radar returns and asynchronous event measurements.

For radar, event cameras, and IMUs, sparse radar returns, asynchronous brightness changes, and inertial measurements do not directly form a shared geometric correspondence. This motivates investigating motion as the linking quantity.

\subsection{Motion-Based Spatiotemporal Calibration}
Motion constraints offer a route to calibration when direct cross-modal correspondence is difficult. Their parameter coverage depends on the motion signal: angular-motion correlation constrains rotation and timing, while translation requires additional information. Wise et al. use radar ego-velocity and camera motion to estimate radar--camera spatial and temporal parameters and visual translation scale \cite{wise2023radarcam}. Their identifiability analysis makes the dependence on motion excitation explicit.

Other methods estimate motion jointly with calibration. RIs-Calib uses continuous-time rotation and ego-velocity for multiple radars and IMUs \cite{riscalib2024}, and iKalibr extends continuous-time targetless calibration to integrated inertial systems \cite{chen2025ikalibr}. MI-Calib estimates multi-IMU calibration using inertial measurements alone \cite{chen2024mi}. Pan et al. represent motion using a multioutput Gaussian process and refine LiDAR point motion for multisensor calibration \cite{pan2026m2calibr}. These methods differ in sensor inputs and motion representations, but all need to account for the interaction between motion and calibration states. PLS-Calib uses motion correlation for event-camera and odometry rotation calibration under ground-motion constraints, with event target detection as part of its formulation \cite{li2026plscalib}. The reference-conditioned construction is related to velocity-domain hand--eye calibration, but here the reference twist enters as a synchronized input to modality-specific residuals rather than as a pose-equation target.

The measurements used to form motion constraints also determine which parameters can be recovered. Multiple IMUs can be calibrated by exploiting differences in their inertial measurements, either with camera assistance or using inertial data alone \cite{rehder2016extending,lee2022extrinsic}. Event--IMU methods use rotational motion derived from events and gyroscopes to estimate rotation and timing \cite{li2025rotation}. For radar--IMU calibration, RadarCalib~\cite{li2026mitigating} extracts rotational information from radar spectral changes rather than relying on point correspondences. These approaches exploit modality-specific information but do not share identical parameter coverage or measurement assumptions.

The recoverable parameters also depend on motion excitation. Analyses of aided inertial navigation and visual--inertial self-calibration identify observability conditions that depend on motion and available measurements \cite{yang2019degenerate,huai2022observability}. OA-LICalib~\cite{lv2022observability} selects informative LiDAR--IMU data and limits updates along weakly observable directions. Other work uses excitation analysis for online LiDAR--IMU calibration \cite{wang2025fly} or studies motion requirements for GNSS lever-arm estimation \cite{wodtko2024globally}. These analyses and safeguards depend on their respective measurement models; the observable parameters must be assessed again for a different sensor suite and reference input.

External motion observations provide another route to alignment. For an omnidirectional vehicle, event-derived and wheel-odometry velocities have been correlated to estimate camera rotation and timing \cite{li2024spatio}. Reference trajectories are also used for spatiotemporal hand--eye alignment \cite{shu2024spatiotemporal}, while raw GNSS observations can aid joint LiDAR--IMU or IMU--odometer calibration \cite{li2023targetless,song2026online}. These examples differ in reference type and whether reference measurements are jointly optimized; they show that external information can complement sensor-specific constraints, but does not automatically eliminate estimation uncertainty.

A fixed reference twist is an alternative to estimating the shared trajectory inside the calibration problem. It permits sensor-specific residual blocks to be evaluated against the same motion and, when no shared parameters couple the blocks, independently refined. The reference itself must be obtained and synchronized, and its errors propagate into calibration. The resulting estimates remain conditional on reference quality and excitation; fixed-reference processing does not eliminate coupling among a sensor's own spatial and temporal parameters. Joint motion-and-calibration methods remain applicable when an external reference is unavailable.

\subsection{Radar/Event/Inertial Ego-Motion Estimation}
Motion estimation is both an application of calibration and a source of constraints for it. We first review single-modality motion estimation and then the cross-modal designs that couple these measurements.

For radar, ego-motion can be inferred from Doppler measurements without point correspondences. Credible velocity estimation and Doppler-based motion recovery show that radial velocity is informative but sensitive to outliers, scene conditions, and sensor geometry \cite{haggag2022credible,galeote2024doppler}. Event cameras provide a complementary motion cue: contrast maximization estimates optical flow, depth, and camera motion \cite{shiba2024secrets}, while line-event geometry supports event-based motion estimation \cite{zhao2025full}. Inertial preintegration provides relative motion and attitude propagation from angular rate and specific force \cite{forster2017preintegration}. These single-modality models establish what each sensor can observe on its own, which is the basis for deciding what must be supplied by calibration or by a reference.

Cross-modal systems couple these cues. Radar--inertial estimators combine Doppler ego-velocity with inertial propagation: River estimates continuous-time rotation and velocity \cite{chen2024river}, iRIOM fuses radar ego-velocity, scan-to-submap constraints, and inertial measurements \cite{zhuang20234d}, and Gaussian-process motion priors support continuous-time radar--inertial and LiDAR--inertial odometry \cite{burnett2024continuous}. Doppler-SLAM incorporates online extrinsic calibration \cite{wang2025doppler}, and temporal calibration has also been integrated into radar--inertial odometry \cite{vstironja2025impact}. Event--inertial systems include ESVIO \cite{chen2023esvio}, instantaneous velocity estimation without a position map \cite{lu2024velometer}, and asynchronous event--inertial motion estimation \cite{wang2025asyneio}; event--IMU calibration methods estimate rotation and timing from rotational motion \cite{li2026iei,mai2025eventcalib}. At the radar--event level, Lyu et al.~\cite{lyu2026radarevent} present, to our knowledge, the first direct event--radar velocity estimator. Their formulation uses the complete rigid-body velocity relation, retaining the lever-arm term, and therefore provides a direct velocity-domain link between event measurements and radar Doppler observations.

These works show that radar, event, and inertial measurements provide motion cues with different parameter coverage: some recover full pose trajectories, others rotation and timing, and others partial velocity. They do not by themselves define a common calibration model for a radar--event--IMU suite. The present work retains the rigid-body lever-arm contribution when relating each sensor origin to a reference twist and treats modality-specific residuals as distinct constraints on the same motion, rather than assuming that the measurements are interchangeable.

\section{From Partial Velocity Observations to Reference-Decoupled Calibration}
\label{sec:reference_decoupling}

This section establishes the observation and estimation model used in the remainder of the paper. It compares joint and reference-conditioned estimation as two probabilistic information structures. The analysis is local and conditional: it provides a design rationale and operating conditions, not an infeasibility proof for reference-free joint estimation or a general superiority claim over it. The reference is treated as a conditioning input, not as a necessary source of observability.

\subsection{Structural Separability and Problem Setting}
\label{subsec:problem_setting}
Let $\mathcal B$ be the calibration body frame and $\mathcal S\in\{\mathcal R,\mathcal E,\mathcal I\}$ index the mmWave radar, event camera, and IMU. The reference supplies the body twist
\begin{equation}
\bm{\xi}_{\mathcal B}(t)=
\begin{bmatrix}\bm{v}_{\mathcal B}(t)\\ \bm{\omega}_{\mathcal B}(t)\end{bmatrix}
\in\mathbb R^6.
\label{eq:body_twist}
\end{equation}
Each sensor reuses the same spatiotemporal state
\begin{equation}
\bm{\theta}_{\mathcal S}^{\mathrm{st}}=
\left\{{}^{\mathcal B}\bm{R}_{\mathcal S},
\;{}^{\mathcal B}\bm{p}_{\mathcal S},
\;\tau_{\mathcal S}\right\},
\label{eq:spatiotemporal_state}
\end{equation}
where ${}^{\mathcal B}\bm{R}_{\mathcal S}$ and ${}^{\mathcal B}\bm{p}_{\mathcal S}$ are the rotation and lever arm from sensor to body, and $\tau_{\mathcal S}$ maps sensor time to reference time. Modality-specific parameters augment this shared state as
\begin{equation}
\begin{aligned}
\bm{\beta}_{\mathcal R}
&=\left[(\bm{b}_{\mathcal R}^{\mathcal B})^{\top},\;s_{\mathcal R}\right]^{\top},
\qquad
\bm{\beta}_{\mathcal E}=\varnothing,\\
\bm{\beta}_{\mathcal I}
&=\left[\bm{b}_g^{\top},\;(\bm{b}_a^{\mathcal B})^{\top},\;s_a\right]^{\top},
\end{aligned}
\label{eq:augmented_params}
\end{equation}
where $\bm{b}_{\mathcal R}^{\mathcal B}$ is a body-frame radar velocity bias, $s_{\mathcal R}$ is the radar velocity scale, $\bm{b}_g$ and $\bm{b}_a^{\mathcal B}$ are gyroscope and body-frame accelerometer biases, and $s_a$ is the accelerometer scale; event-camera intrinsics are calibrated separately. The augmented sensor state is $\bm{\theta}_{\mathcal S}=(\bm{\theta}_{\mathcal S}^{\mathrm{st}},\bm{\beta}_{\mathcal S})$. To keep the notation compact, body quantities in the $\mathcal S$-th model are evaluated at the aligned time $t_k-\tau_{\mathcal S}$.

For a rigid mount, the sensor-origin velocity is
\begin{equation}
\begin{bmatrix}
{}^{\mathcal S}\bm{v}_{\mathcal S,k}\\
{}^{\mathcal S}\bm{\omega}_{\mathcal S,k}
\end{bmatrix}
=
\begin{bmatrix}
\bm{R}_{\mathcal S\mathcal B}
\left(\bm{v}_{\mathcal B,k}+\bm{\omega}_{\mathcal B,k}\times\bm{p}_{\mathcal S}\right)\\
\bm{R}_{\mathcal S\mathcal B}\bm{\omega}_{\mathcal B,k}
\end{bmatrix},
\label{eq:lever}
\end{equation}
where $\bm{R}_{\mathcal S\mathcal B}=({}^{\mathcal B}\bm{R}_{\mathcal S})^\top$ and $\bm{p}_{\mathcal S}={}^{\mathcal B}\bm{p}_{\mathcal S}$. The first row is the linear velocity at the sensor origin and the second row is the angular velocity. Translation enters only through the lever-arm velocity and is therefore observable through rotational excitation.

\subsubsection{Radar Ego-Velocity Observation}
Following the radar-inertial velocity formulation in River~\cite{chen2024river}, the radar observation is
\begin{equation}
\bm{z}_{\mathcal R,k}=
\bm{R}_{\mathcal R\mathcal B}\left[
s_{\mathcal R}\left(\bm{v}_{\mathcal B,k}+\bm{\omega}_{\mathcal B,k}\times\bm{p}_{\mathcal R}\right)
+\bm{b}_{\mathcal R}^{\mathcal B}\right]
+\bm{\epsilon}_{\mathcal R,k},
\label{eq:radar_observation}
\end{equation}
where $\bm{z}_{\mathcal R,k}\in\mathbb R^3$ is the radar ego-velocity. The bias $\bm{b}_{\mathcal R}^{\mathcal B}$ and scale $s_{\mathcal R}$ absorb a constant body-frame velocity offset and a reference-to-radar velocity scale, respectively. Radar has no intrinsic angular-velocity channel; its angular sensitivity is produced solely by $\bm{p}_{\mathcal R}$.

\subsubsection{Event Normal-Flow Observation}
Following the radar--event velocity relation of Lyu et al.~\cite{lyu2026radarevent}, let $\bm{z}_{\mathcal E,k}=(\bm{x}_k,\bm{n}_k,v_{n,k})$ collect the normalized image point, unit gradient direction, and normal-flow magnitude. The scalar normal-flow observation satisfies
\begin{equation}
0=
\left([\bm{x}_k]_\times{}^{\mathcal E}\bm{v}_{\mathcal E,k}\right)^\top
\left(v_{n,k}\bm{n}_k+
[\bm{x}_k]_\times{}^{\mathcal E}\bm{\omega}_{\mathcal E,k}\right)
+\epsilon_{\mathcal E,k}.
\label{eq:event_observation}
\end{equation}
Event normal flow is therefore a scalar projection of image-plane motion, not a vector velocity measurement, and it is bilinear in linear and angular velocity.

\subsubsection{Inertial Observation}
Using the standard IMU model~\cite{forster2017preintegration}, define the reference kinematic acceleration at the body origin $\bm{u}_{r,k}=\dot{\bm{v}}_{\mathcal B,k}+\bm{\omega}_{\mathcal B,k}\times\bm{v}_{\mathcal B,k}$ and $\bm{L}_k=[\dot{\bm{\omega}}_{\mathcal B,k}]_\times+[\bm{\omega}_{\mathcal B,k}]_\times^2$. The gyroscope and accelerometer observations are
\begin{equation}
\begin{aligned}
\bm{z}^{\omega}_{\mathcal I,k}&=
{}^{\mathcal I}\bm{\omega}_{\mathcal I,k}+\bm{b}_g+\bm{\epsilon}_{g,k},\\
\bm{z}^{a}_{\mathcal I,k}&=
\bm{R}_{\mathcal I\mathcal B}\!\left[
s_a\left(\bm{u}_{r,k}+\bm{L}_k\bm{p}_{\mathcal I}-\bm{g}_{\mathcal B,k}\right)
+\bm{b}_a^{\mathcal B}\right]+\bm{\epsilon}_{a,k}.
\end{aligned}
\label{eq:imu_observation}
\end{equation}
The gyroscope provides direct angular-rate information; the accelerometer provides a gravity-containing specific force rather than a linear-velocity measurement.

\subsubsection{Structural Separability}
\begin{figure*}[t]
\centering
\tikzset{
  obstri/.style={regular polygon,regular polygon sides=3,draw=black,line width=0.55pt,minimum size=8mm,inner sep=0pt,align=center,font=\fontsize{4.9}{5.5}\selectfont},
  obsR/.style={obstri,fill=green!62},
  obsE/.style={obstri,fill=green!62},
  obsI/.style={obstri,fill=green!62},
  refobs/.style={obstri,fill=green!62},
  factor/.style={circle,draw=black,line width=0.6pt,fill=black!85,text=white,minimum size=8mm,inner sep=0pt,align=center,font=\fontsize{5.0}{5.5}\selectfont},
  statebase/.style={circle,draw=black,dashed,line width=0.8pt,minimum size=8mm,inner sep=0pt,align=center,font=\fontsize{5.4}{6.2}\selectfont},
  theta/.style={statebase,fill=yellow!68},
  thetaB/.style={theta,solid},
  est/.style={statebase,fill=blue!68!black,text=white},
  obsedge/.style={->,draw=black!75,line width=0.55pt},
  paramedge/.style={->,draw=black!75,line width=0.55pt},
  pairedge/.style={draw=black!55,dashed,line width=0.45pt},
  pathA/.style={->,draw=blue!88!black,line width=1.0pt},
  pathB/.style={->,draw=red!80!black,line width=1.1pt},
  twistframe/.style={draw=blue!88!black,dashed,line width=0.7pt,rounded corners=2pt},
  refframe/.style={draw=green!45!black,line width=0.7pt,fill=green!8,rounded corners=1.5pt}
}
\begin{subfigure}[t]{0.49\textwidth}
\centering
\resizebox{\linewidth}{!}{%
\begin{tikzpicture}[font=\scriptsize,>=Stealth,x=0.78cm,y=0.90cm]
\node[obsR] (zR) at (0,2.62) {$\bm{z}_{\mathcal R}$};
\node[obsE] (zE) at (-4.44,-1.05) {$\bm{z}_{\mathcal E}$};
\node[obsI] (zIw) at (4.47,-0.43) {$\bm{z}_{\mathcal I}^{\omega}$};
\node[obsI] (zIa) at (4.24,-1.67) {$\bm{z}_{\mathcal I}^{a}$};
\node[factor] (fR) at (0,1.62) {$f_{\mathcal R}$};
\node[factor] (fE) at (-3.32,-1.05) {$f_{\mathcal E}$};
\node[factor] (fI) at (3.32,-1.05) {$f_{\mathcal I}$};
\node[theta] (tR) at (-1.45,1.12) {$\bm{\theta}_{\mathcal R}$};
\node[theta] (tE) at (-2.20,-1.05) {$\bm{\theta}_{\mathcal E}$};
\node[theta] (tI) at (2.20,-1.05) {$\bm{\theta}_{\mathcal I}$};
\node[est] (vB) at (-0.58,-0.02) {$\bm{v}_{\mathcal B}$};
\node[est] (wB) at (0.58,-0.02) {$\bm{\omega}_{\mathcal B}$};
\draw[twistframe] (-1.28,-0.55) rectangle (1.28,0.50);
\draw[pairedge] (tR.south west)--node[fill=white,inner sep=0.6pt,font=\fontsize{4.5}{5}\selectfont,pos=0.48,left=1pt]{$T_{\mathcal R\mathcal E}$}(tE.north west);
\draw[pairedge] (tE.south east)--node[fill=white,inner sep=0.6pt,font=\fontsize{4.5}{5}\selectfont,pos=0.50,below]{$T_{\mathcal E\mathcal I}$}(tI.south west);
\draw[pairedge] (tR.east) to[out=10,in=145] node[fill=white,inner sep=0.6pt,font=\fontsize{4.5}{5}\selectfont,pos=0.58,above=1pt]{$T_{\mathcal R\mathcal I}$} (tI.north);
\draw[obsedge] (zR)--(fR);
\draw[obsedge] (zE)--(fE);
\draw[obsedge] (zIw)--(fI);
\draw[obsedge] (zIa)--(fI);
\draw[pathA] (fR.south)--(0,0.50);
\draw[pathA] (fE.east)--(-1.28,-0.18);
\draw[pathA] (fI.west)--(1.28,-0.18);
\draw[pathA] (-0.60,0.50)--(tR.south);
\draw[pathA] (-1.28,-0.20)--(tE.east);
\draw[pathA] (1.28,-0.20)--(tI.west);
\end{tikzpicture}%
}
\caption{Joint estimation with a shared latent twist.}
\end{subfigure}
\hfill
\begin{subfigure}[t]{0.49\textwidth}
\centering
\resizebox{\linewidth}{!}{%
\begin{tikzpicture}[font=\scriptsize,>=Stealth,x=0.78cm,y=0.90cm]
\node[obsR] (zR) at (0,2.62) {$\bm{z}_{\mathcal R}$};
\node[obsE] (zE) at (-4.44,-1.05) {$\bm{z}_{\mathcal E}$};
\node[obsI] (zIw) at (4.47,-0.43) {$\bm{z}_{\mathcal I}^{\omega}$};
\node[obsI] (zIa) at (4.24,-1.67) {$\bm{z}_{\mathcal I}^{a}$};
\node[factor] (fR) at (0,1.62) {$f_{\mathcal R}$};
\node[factor] (fE) at (-3.32,-1.05) {$f_{\mathcal E}$};
\node[factor] (fI) at (3.32,-1.05) {$f_{\mathcal I}$};
\node[thetaB] (tR) at (-1.45,1.12) {$\bm{\theta}_{\mathcal R}$};
\node[thetaB] (tE) at (-2.20,-1.05) {$\bm{\theta}_{\mathcal E}$};
\node[thetaB] (tI) at (2.20,-1.05) {$\bm{\theta}_{\mathcal I}$};
\draw[refframe] (-1.45,-0.55) rectangle (1.45,0.50);
\node[refobs] (vR) at (-0.82,-0.02) {$\bm{v}_{\mathrm{ref}}$};
\node[refobs] (wR) at (0.82,-0.02) {$\bm{\omega}_{\mathrm{ref}}$};
\draw[pairedge] (tR.south west)--node[fill=white,inner sep=0.6pt,font=\fontsize{4.5}{5}\selectfont,pos=0.48,left=1pt]{$T_{\mathcal R\mathcal E}$}(tE.north west);
\draw[pairedge] (tE.south east)--node[fill=white,inner sep=0.6pt,font=\fontsize{4.5}{5}\selectfont,pos=0.50,below]{$T_{\mathcal E\mathcal I}$}(tI.south west);
\draw[pairedge] (tR.east) to[out=10,in=145] node[fill=white,inner sep=0.6pt,font=\fontsize{4.5}{5}\selectfont,pos=0.58,above=1pt]{$T_{\mathcal R\mathcal I}$} (tI.north);
\draw[obsedge] (zR)--(fR);
\draw[obsedge] (zE)--(fE);
\draw[obsedge] (zIw)--(fI);
\draw[obsedge] (zIa)--(fI);
\draw[paramedge] (fR)--(tR);
\draw[paramedge] (fE)--(tE);
\draw[paramedge] (fI)--(tI);
\draw[pathB] (-0.82,0.50)--(tR.south);
\draw[pathB] (-1.45,-0.20)--(tE.east);
\draw[pathB] (1.45,-0.20)--(tI.west);
\end{tikzpicture}%
}
\caption{Reference-conditioned sensor-wise estimation.}
\end{subfigure}
\caption{Factor graphs of the two pathways. Green triangles denote observations and reference inputs; circles denote parameters and states; black filled circles are residual factors. Dashed outlines mark states estimated through the shared latent twist, whereas solid outlines mark reference-conditioned parameters. Dashed links denote pairwise relative transforms implied by the sensor states, not independent unknowns.}
\label{fig:reference_graph}
\end{figure*}
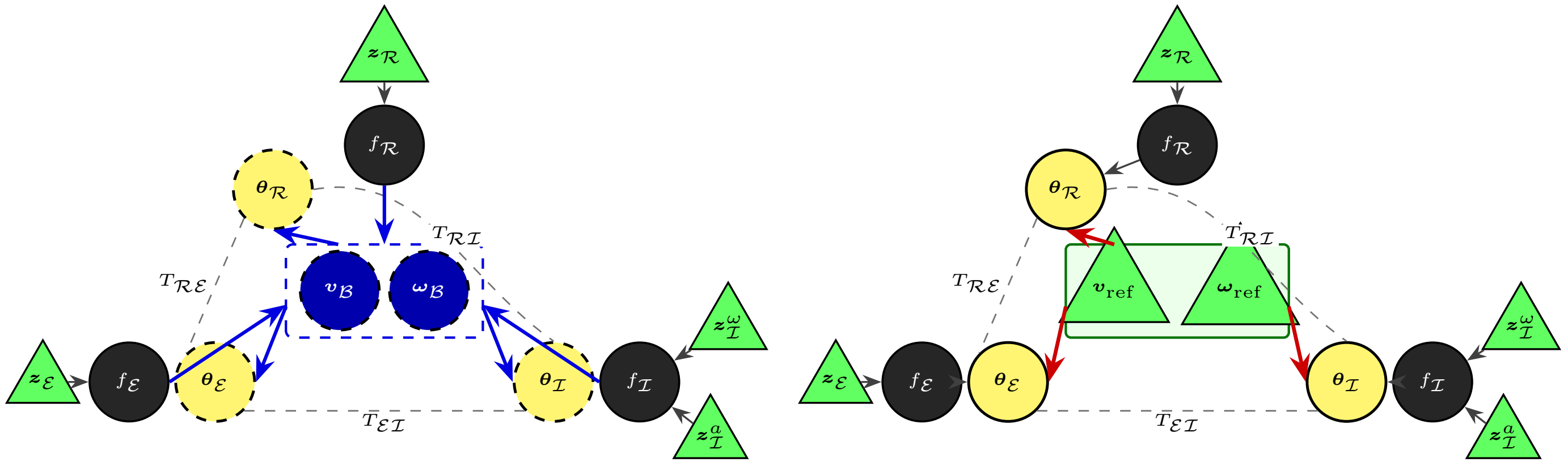

Equations~\eqref{eq:radar_observation}--\eqref{eq:imu_observation} show that every sensor observes a projection or derivative of the common twist. Let $\mathcal K_{\mathcal S}$ index the measurements of sensor $\mathcal S$, and define $\bm{Z}_{\mathcal S}=\{\bm{z}_{\mathcal S,k}:k\in\mathcal K_{\mathcal S}\}$ and $\bm{Z}=\{\bm{Z}_{\mathcal R},\bm{Z}_{\mathcal E},\bm{Z}_{\mathcal I}\}$. For $\mathcal S\ne\mathcal S'$, the $\mathcal S$-th model contains no parameter of sensor $\mathcal S'$, and the noises are independent. Hence the likelihood factorizes as
\begin{equation}
p(\bm{Z}\,|\,\bm{\xi}_{\mathcal B},\{\bm{\theta}_{\mathcal S}\})
=\prod_{\mathcal S}p_{\mathcal S}(\bm{Z}_{\mathcal S}\,|\,\bm{\xi}_{\mathcal B},\bm{\theta}_{\mathcal S}),
\label{eq:separable_likelihood}
\end{equation}
where $p_{\mathcal S}$ denotes the $\mathcal S$-th sensor likelihood factor. The sensor blocks are \emph{conditionally independent given the shared twist}. Two consequences follow. First, $\bm{Z}_{\mathcal S}$ contains no direct information about another sensor's parameters: all cross-sensor parameter coupling must pass through $\bm{\xi}_{\mathcal B}$, so there is no direct cross-modal residual. Second, no single modality supplies a complete body twist under \eqref{eq:radar_observation}--\eqref{eq:imu_observation}. Radar has no intrinsic angular-velocity channel, event normal flow is a scalar projection, and the IMU does not directly measure linear velocity. The blocks are therefore informationally isolated and individually partial; the shared twist is their only common endogenous state when no external reference is supplied, as shown in Fig.~\ref{fig:reference_graph}. Consequently, the blocks must share either an estimated twist or an external reference, and neither option creates missing native sensitivity.

\subsection{Estimation Structure}
\label{subsec:estimation_structure}
Let $\bm{\theta}=\{\bm{\theta}_{\mathcal S}\}$ collect all sensor parameters and let $n_\theta$ be its dimension. Pathway A treats the shared twist as a latent state and estimates it jointly with the sensor parameters, whereas Pathway B conditions on a trusted reference:
\begin{equation}
p_A(\{\bm{\theta}_{\mathcal S}\},\bm{\xi}_{\mathcal B}\,|\,\bm{Z})
\propto
\left[\prod_{\mathcal S}p_{\mathcal S}(\bm{Z}_{\mathcal S}\,|\,\bm{\theta}_{\mathcal S},\bm{\xi}_{\mathcal B})p(\bm{\theta}_{\mathcal S})\right]
p(\bm{\xi}_{\mathcal B}),
\label{eq:joint_posterior}
\end{equation}
\begin{equation}
p_B(\{\bm{\theta}_{\mathcal S}\}\,|\,\bm{Z},\bm{\xi}_{\mathrm{ref}})
\propto
\prod_{\mathcal S}p_{\mathcal S}(\bm{Z}_{\mathcal S}\,|\,\bm{\theta}_{\mathcal S},\bm{\xi}_{\mathrm{ref}})p(\bm{\theta}_{\mathcal S}).
\label{eq:conditional_posterior}
\end{equation}
The factorization follows from \eqref{eq:separable_likelihood}, not from parameter independence. In \eqref{eq:joint_posterior}, the latent twist couples all blocks; in \eqref{eq:conditional_posterior}, fixing $\bm{\xi}_{\mathrm{ref}}$ yields sensor-wise subproblems, with $p(\bm{\theta}_{\mathcal S})$ the block prior or regularizer (constant when flat). Let $\bm{\xi}_{\mathcal B}^{\star}$ denote the true body twist and $\delta\bm{\xi}_{\mathrm{ref}}=\bm{\xi}_{\mathrm{ref}}-\bm{\xi}_{\mathcal B}^{\star}$ the reference error. The reference therefore breaks the information isolation without adding a likelihood term for $\bm{\theta}_{\mathcal S}$.

Linearizing the stacked residual $\bm{r}$ around a local solution gives $\delta\bm{r}\approx\bm{A}\delta\bm{\xi}_{\mathcal B}+\bm{B}\delta\bm{\theta}$, with $\bm{A}=\partial\bm{r}/\partial\bm{\xi}_{\mathcal B}$, $\bm{B}=\operatorname{blkdiag}(\bm{B}_{\mathcal R},\bm{B}_{\mathcal E},\bm{B}_{\mathcal I})$, $\bm{B}_{\mathcal S}=\partial\bm{r}_{\mathcal S}/\partial\bm{\theta}_{\mathcal S}$, and $\bm{W}=\operatorname{blkdiag}(\bm{W}_{\mathcal R},\bm{W}_{\mathcal E},\bm{W}_{\mathcal I})$. The information blocks are
\begin{equation}
\begin{aligned}
\bm{F}_{\theta\theta}&=\bm{B}^{\top}\bm{W}\bm{B}
=\operatorname{blkdiag}(\bm{F}_{\mathcal R},\bm{F}_{\mathcal E},\bm{F}_{\mathcal I}),\\
\bm{F}_{\xi\xi}&=\bm{A}^{\top}\bm{W}\bm{A},\qquad
\bm{F}_{\xi\theta}=\bm{A}^{\top}\bm{W}\bm{B},\qquad
\bm{F}_{\theta\xi}=\bm{F}_{\xi\theta}^{\top}.
\end{aligned}
\label{eq:information_blocks}
\end{equation}
Eliminating $\delta\bm{\xi}_{\mathcal B}$ from Pathway A by the Schur complement gives its parameter-only normal equation. Pathway B already has this form:
\begin{equation}
\begin{aligned}
\text{A:}\quad&
\bm{S}_A\,\delta\bm{\theta}
=\bm{g}_\theta-\bm{F}_{\theta\xi}\bm{F}_{\xi\xi}^{-1}\bm{g}_\xi,\\
& \bm{S}_A=\bm{F}_{\theta\theta}-\bm{F}_{\theta\xi}\bm{F}_{\xi\xi}^{-1}\bm{F}_{\xi\theta},\\
\text{B:}\quad&
\bm{F}_{\theta\theta}\,\delta\bm{\theta}
=\bm{g}_\theta-\bm{F}_{\theta\xi}\,\delta\bm{\xi}_{\mathrm{ref}},
\end{aligned}
\label{eq:schur_normal_equations}
\end{equation}
where $\bm{g}_\theta$ and $\bm{g}_\xi$ are the local gradients in the two blocks. The two rows are alternative normal equations for the same $\delta\bm{\theta}$ and are not solved simultaneously: the first eliminates the shared twist, while the second uses the reference as a known input. Pathway A uses the dense Schur-reduced $\bm{S}_A$ and the state-elimination correction $-\bm{F}_{\theta\xi}\bm{F}_{\xi\xi}^{-1}\bm{g}_\xi$; Pathway B uses the block-diagonal $\bm{F}_{\theta\theta}$ and the external correction $-\bm{F}_{\theta\xi}\delta\bm{\xi}_{\mathrm{ref}}$. A carries the shared-state bottleneck, whereas B removes it from the normal equations at the cost of a common exogenous reference error. If $\bm{F}_{\theta\xi}=\bm{0}_{n_\theta\times6}$, the two coincide up to the conditioning correction.

The reference does not make the blocks statistically independent, because its error is common; nor does it enlarge $\bm{F}_{\theta\theta}$ or recover directions absent from native observations.

\subsection{Identifiability, Conditioning, and Error Propagation}
\label{subsec:schur_conditioning}
After gauge fixing, let $\bm{\Sigma}_{\xi,\mathrm{ref}}$ be the reference covariance and $\bm{\Lambda}_{\mathrm{ref}}=\bm{\Sigma}_{\xi,\mathrm{ref}}^{-1}$ its information matrix, with $\bm{\Lambda}_{\mathrm{ref}}=\bm{0}_{6\times6}$ for no reference and $\|\bm{\Lambda}_{\mathrm{ref}}\|\to\infty$ for an exact reference. Eliminating the shared twist gives
\begin{equation}
\bm{S}(\bm{\Lambda}_{\mathrm{ref}})=
\bm{F}_{\theta\theta}-
\bm{F}_{\theta\xi}
\left(\bm{F}_{\xi\xi}+\bm{\Lambda}_{\mathrm{ref}}\right)^{-1}
\bm{F}_{\xi\theta},
\label{eq:schur}
\end{equation}
with $\bm{F}_{\xi\xi}+\bm{\Lambda}_{\mathrm{ref}}\succ\bm{0}_{6\times6}$. The Schur penalty is
\begin{equation}
\Delta\bm{S}(\bm{\Lambda}_{\mathrm{ref}})
=\bm{F}_{\theta\theta}-\bm{S}(\bm{\Lambda}_{\mathrm{ref}})
=\bm{F}_{\theta\xi}
\left(\bm{F}_{\xi\xi}+\bm{\Lambda}_{\mathrm{ref}}\right)^{-1}
\bm{F}_{\xi\theta}.
\label{eq:schur_penalty}
\end{equation}
Pathway A, with $\bm{S}_A=\bm{S}(\bm{0}_{6\times6})$, pays the full penalty $\bm{F}_{\theta\xi}\bm{F}_{\xi\xi}^{-1}\bm{F}_{\xi\theta}$; an exact reference removes it, and a finite reference removes part of it. Thus A can be singular or ill-conditioned through the Schur subtraction even when $\bm{F}_{\theta\theta}$ is nonsingular, whereas B is rank-limited only by $\bm{F}_{\theta\theta}$. This is the conditioning difference between the two normal equations in \eqref{eq:schur_normal_equations}.

The covariance and first-order bias are
\begin{equation}
\begin{aligned}
\bm{\Sigma}_\theta(\bm{\Lambda}_{\mathrm{ref}})
&=\bm{S}(\bm{\Lambda}_{\mathrm{ref}})^{-1},\\
\bm{b}_\theta(\bm{\Lambda}_{\mathrm{ref}})
&=
-\bm{S}(\bm{\Lambda}_{\mathrm{ref}})^{-1}
\bm{F}_{\theta\xi}
\left(\bm{F}_{\xi\xi}+\bm{\Lambda}_{\mathrm{ref}}\right)^{-1}
\bm{\Lambda}_{\mathrm{ref}}\,\delta\bm{\xi}_{\mathrm{ref}},
\end{aligned}
\label{eq:covariance_bias}
\end{equation}
Pathway A has covariance $\bm{\Sigma}_\theta(\bm{0}_{6\times6})=\bm{S}_A^{-1}$; an exact-reference B has covariance $\bm{F}_{\theta\theta}^{-1}$ and bias $-\bm{F}_{\theta\theta}^{-1}\bm{F}_{\theta\xi}\delta\bm{\xi}_{\mathrm{ref}}$. For a finite reference, the covariance lies between these limits, whereas the bias scales with $\bm{\Lambda}_{\mathrm{ref}}$ toward the exact-reference value. The covariance gain is structural, but the reference introduces a bias term that A does not contain.

The monotonicity $\bm{S}_A\preceq\bm{S}(\bm{\Lambda}_{\mathrm{ref}})\preceq\bm{F}_{\theta\theta}$ and inversion of the Loewner order give
\begin{equation}
\bm{\Sigma}_\theta(\infty)
\preceq\bm{\Sigma}_\theta(\bm{\Lambda}_{\mathrm{ref}})
\preceq\bm{\Sigma}_\theta(\bm{0}_{6\times6}),
\label{eq:covariance_order}
\end{equation}
where $\bm{\Sigma}_\theta(\infty)$ denotes the exact-reference limit. For a locally linear estimator, let $\bm{M}_\theta(\bm{\Lambda}_{\mathrm{ref}})=\bm{\Sigma}_\theta(\bm{\Lambda}_{\mathrm{ref}})+\bm{b}_\theta(\bm{\Lambda}_{\mathrm{ref}})\bm{b}_\theta(\bm{\Lambda}_{\mathrm{ref}})^{\top}$ be the mean-squared-error (MSE) matrix. For an unbiased joint estimator, $\bm{M}_\theta(\bm{0}_{6\times6})=\bm{\Sigma}_\theta(\bm{0}_{6\times6})$.

\textbf{Proposition 1 (CRLB-level structural gain).} Suppose the local blocks are identifiable, the reference is independent and synchronized, and the bias satisfies
\begin{equation}
\|\bm{b}_\theta(\bm{\Lambda}_{\mathrm{ref}})\|^2\le
\lambda_{\min}\!\left(\bm{\Sigma}_\theta(\bm{0}_{6\times6})-\bm{\Sigma}_\theta(\bm{\Lambda}_{\mathrm{ref}})\right).
\label{eq:bias_bound}
\end{equation}
Then
\begin{equation}
\bm{M}_\theta(\bm{\Lambda}_{\mathrm{ref}})
\preceq\bm{\Sigma}_\theta(\bm{0}_{6\times6}),
\label{eq:mse_bound}
\end{equation}
with strict inequality when \eqref{eq:bias_bound} is strict. This is a CRLB-level statement about information structures, not a convergence guarantee for a particular nonlinear solver.

The reference does not add physical observations. If $\bm{F}_{\theta\theta}\bm{v}=\bm{0}_{n_\theta\times1}$, then $\bm{S}(\bm{\Lambda}_{\mathrm{ref}})\bm{v}=\bm{0}_{n_\theta\times1}$ for every reference, so both pathways remain unobservable in that native direction. Otherwise, Pathway B removes the Schur loss but carries the reference bias, while Pathway A avoids that bias but carries the shared-state bottleneck. Global identifiability additionally requires sustained excitation over the calibration window.

\subsection{Structural Conclusion and Operating Conditions}
\label{subsec:operating_conditions}
At this local CRLB level, Pathway B uses the block information $\bm{F}_{\theta\theta}$ rather than the joint marginal $\bm{S}_A$. Since $\bm{S}_A\preceq\bm{F}_{\theta\theta}$, its covariance cannot be larger for any linear functional of $\bm{\theta}$, and it can gain on nuisance-coupled weak directions. The gain is structural and requires an independent, synchronized, sufficiently excited reference; finite noise reduces it and introduces bias. Reference noise does not create a normal-equation feedback term, but it is injected commonly into all blocks, so decoupling is structural rather than statistical. A timing error acts through $\dot{\bm{\xi}}_{\mathrm{ref}}$ and can mimic lever-arm translation, so synchronization cannot be replaced by decoupling. Pathway A remains valid when no trusted reference is available, and the experiments test these operating conditions.

\section{Method}
This section instantiates the model of Section~\ref{sec:reference_decoupling} for each modality. It details residual construction, initialization, and block refinement; the estimation-structure and observability analysis is not repeated. For each sensor, the parameters are obtained from
\begin{equation}
\hat{\bm{\theta}}_{\mathcal S}
=\arg\min_{\bm{\theta}_{\mathcal S}}\sum_k
\rho\!\left(\left\|\bm{r}_{\mathcal S,k}(\bm{\theta}_{\mathcal S})\right\|^2\right),
\label{eq:general_objective}
\end{equation}
where $\bm{r}_{\mathcal S,k}$ is the residual induced by the corresponding observation in \eqref{eq:radar_observation}--\eqref{eq:imu_observation}, and $\rho(\cdot)$ is a robust loss (Huber in the implementation).

\begin{figure*}[t]
\centering
\includegraphics[width=\textwidth]{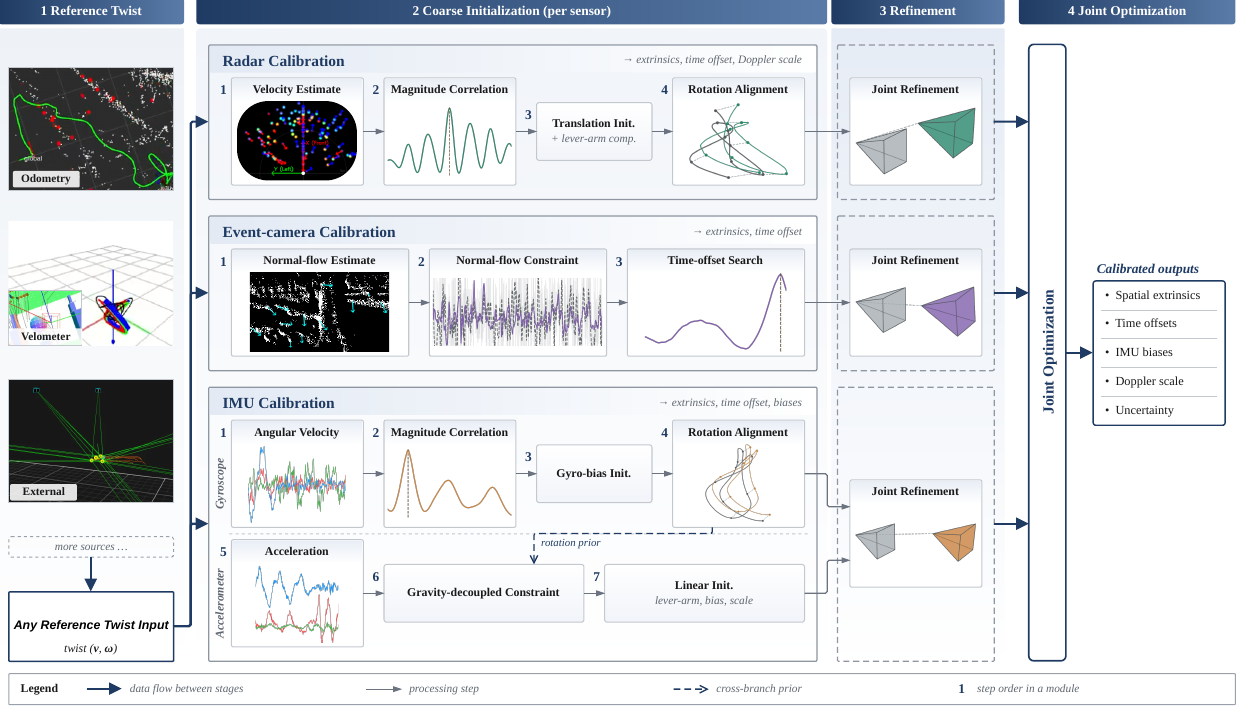}
\caption{Technical framework. Motion cues from radar Doppler, event normal flow, and IMU gyroscope are aligned to a reference twist. Time offsets are initialized by motion-cue correlation, then spatial and temporal parameters are optimized through velocity-domain residuals. Outputs include radar/event/IMU extrinsics, time offsets, uncertainty, residual diagnostics, and downstream configuration files.}
\label{fig:framework}
\end{figure*}

\subsection{Radar Calibration}
\label{sec:radar_calibration}

\subsubsection{Initial Guess}
\label{sec:radar_initial_guess}

Let $\bm{u}_{\mathcal R,k}=\bm{v}_{\mathcal B,k}+\bm{\omega}_{\mathcal B,k}\times\bm{p}_{\mathcal R}$. With the bias set to zero for initialization, taking the norm of \eqref{eq:radar_observation} eliminates the rotation:
\begin{equation}
\left\|\bm{z}_{\mathcal{R},k}\right\|=
s_{\mathcal R}\left\|\bm{u}_{\mathcal{R},k}\right\|,
\label{eq:radar_magnitude}
\end{equation}
because $\bm{R}_{\mathcal{R}\mathcal{B}}$ is orthogonal. Initialization proceeds in three steps. We first approximate the lever arm as negligible and estimate $\tau_{\mathcal R}^0$; with $\tau_{\mathcal R}^0$ fixed, we then initialize the lever arm and finally the rotation. The lever arm is therefore omitted only in the first step and is restored before rotation estimation.

When rotation is weak, the lever-arm term is negligible and \eqref{eq:radar_magnitude} reduces to $\|\bm{z}_{\mathcal{R},k}\|\approx s_{\mathcal R}\|\bm{v}_{\mathcal{B},k}\|$. The time offset is initialized by maximizing the normalized magnitude correlation,
\begin{equation}
\tau_{\mathcal R}^0 = \arg\max_{\tau'} \;
\frac{
\sum_k \|\bm{v}_{\mathcal{B}}(t_k)\| \,
\|\bm{z}_{\mathcal{R}}(t_k + \tau')\|
}{
\sqrt{\sum_k \|\bm{v}_{\mathcal{B}}(t_k)\|^2} \,
\sqrt{\sum_k \|\bm{z}_{\mathcal{R}}(t_k + \tau')\|^2}
},
\label{eq:cross_correlation}
\end{equation}
where the constant radar velocity scale cancels out.

To retain the lever-arm signal under general motion, subtract $s_{\mathcal R}^2\|\bm{v}_{\mathcal{B},k}\|^2$ from the squared norm:
\begin{equation}
\Delta_k =
2s_{\mathcal R}^2\left(\bm{v}_{\mathcal{B},k}\times\bm{\omega}_{\mathcal{B},k}\right)^{\top}\bm{p}_{\mathcal{R}}
+s_{\mathcal R}^2\left\|\bm{\omega}_{\mathcal{B},k}\times\bm{p}_{\mathcal{R}}\right\|^2,
\label{eq:radar_delta}
\end{equation}
where $\Delta_k=\|\bm{z}_{\mathcal{R},k}\|^2-s_{\mathcal R}^2\|\bm{v}_{\mathcal{B},k}\|^2$. With $s_{\mathcal R}=1$, $\bm{b}_{\mathcal R}^{\mathcal B}=\bm{0}_{3\times1}$, and $\tau_{\mathcal R}^0$ fixed, neglecting the quadratic term gives the linear initialization
\begin{equation}
\bm{p}_{\mathcal{R}}^0 = 
\arg\min_{\bm{p}_{\mathcal{R}}}
\sum_{k=1}^{N}
\left(\Delta_k-
2\left(\bm{v}_{\mathcal{B},k}\times\bm{\omega}_{\mathcal{B},k}\right)^{\top}
\bm{p}_{\mathcal{R}}\right)^2.
\label{eq:translation_linear}
\end{equation}
The quadratic term is then restored by Gauss--Newton refinement. Under single-axis rotation or weak excitation, unobservable components are not reported. With $\tau_{\mathcal R}^0$ and $\bm{p}_{\mathcal R}^0$ initialized, and $\bm{u}_{\mathcal R,k}$ evaluated at $\bm{p}_{\mathcal R}^0$, the rotation follows from the Wahba problem
\begin{equation}
{}^{\mathcal{B}}\bm{R}_{\mathcal{R}}^0 = 
\arg\min_{\bm{R} \in \mathrm{SO}(3)} 
\sum_{k \in \mathcal{J}_{\mathcal R}} \left\|\bm{u}_{\mathcal{R},k}-
\bm{R}\,\bm{z}_{\mathcal{R},k}\right\|^2,
\label{eq:wahba}
\end{equation}
where $s_{\mathcal R}=1$, $\bm{b}_{\mathcal R}^{\mathcal B}=\bm{0}_{3\times1}$, and the inlier set $\mathcal{J}_{\mathcal R}$ is obtained by RANSAC. Equation~\eqref{eq:wahba} is a Wahba problem~\cite{wahba1965}, solved in closed form by the Kabsch algorithm~\cite{kabsch1976}; the result initializes refinement.

\subsubsection{Refinement}
\label{sec:radar_refinement}

Starting from the initial guess, all radar parameters are refined with the residual induced by \eqref{eq:radar_observation}. The bias is initialized to zero and refined jointly with the extrinsic and temporal states. The velocity scale is initialized from the magnitude ratio and parameterized as $s_{\mathcal R}=\exp(\alpha_{\mathcal R})$ to enforce positivity. The robust objective \eqref{eq:general_objective} is solved by Levenberg--Marquardt on $\mathrm{SO}(3)$; the loss suppresses moving-object and multipath outliers.

\subsection{Event Camera Calibration}
\label{sec:event_calibration}

The event-camera intrinsics are calibrated separately, so the normalized image point $\bm{x}_k$ in \eqref{eq:event_observation} is known. The scalar event residual $r_{\mathcal E,k}$ is the left-hand side of \eqref{eq:event_observation} without $\epsilon_{\mathcal E,k}$. It directly couples rotation, lever arm, and time offset through the normal-flow measurement.

Because this residual is non-convex in $\bm{\theta}_{\mathcal E}$, we initialize $\tau_{\mathcal E}$ by a coarse grid search and set the small-lever approximation $\bm{p}_{\mathcal E}=\bm{0}_{3\times1}$ before solving \eqref{eq:general_objective}. We use Levenberg--Marquardt on $\mathrm{SO}(3)$ with a Huber loss to suppress dynamic-object and normal-flow outliers.

\subsection{IMU Calibration}
\label{sec:imu_calibration}

The IMU state $\bm{\theta}_{\mathcal I}$ is defined by \eqref{eq:spatiotemporal_state} and \eqref{eq:augmented_params}. The gyroscope and accelerometer stages are solved sequentially; gravity is not an optimization variable.

\subsubsection{Gyroscope Initialization}
\label{sec:imu_gyro_init}

Taking the squared norm of the gyroscope model in \eqref{eq:imu_observation} gives
\begin{equation}
\left\|\bm{z}^{\omega}_{\mathcal{I},k}\right\|^2=
\left\|\bm{\omega}_{\mathcal{B},k}\right\|^2
+2\bm{\omega}_{\mathcal{B},k}^{\top}\bm{c}
+\left\|\bm{c}\right\|^2,
\label{eq:gyro_magnitude}
\end{equation}
where $\bm{c}={}^{\mathcal{B}}\bm{R}_{\mathcal{I}}\bm{b}_g$ is the gyroscope bias expressed in the body frame. Since the bias is small, its coupling is negligible for initialization. The time offset is initialized by maximizing the normalized magnitude correlation,
\begin{equation}
\tau_{\mathcal{I}}^0 = \arg\max_{\tau'} \;
\frac{
\sum_k \|\bm{\omega}_{\mathcal{B}}(t_k)\| \,
\|\bm{z}^{\omega}_{\mathcal{I}}(t_k + \tau')\|
}{
\sqrt{\sum_k \|\bm{\omega}_{\mathcal{B}}(t_k)\|^2} \,
\sqrt{\sum_k \|\bm{z}^{\omega}_{\mathcal{I}}(t_k + \tau')\|^2}
}.
\label{eq:gyro_cross_correlation}
\end{equation}
With $\tau_{\mathcal{I}}^0$ fixed, the difference of squared magnitudes gives
\begin{equation}
\Delta_{\omega,k} =
\left\|\bm{z}^{\omega}_{\mathcal{I},k}\right\|^2-
\left\|\bm{\omega}_{\mathcal{B},k}\right\|^2
\approx 2\bm{\omega}_{\mathcal{B},k}^{\top}\bm{c},
\label{eq:gyro_delta}
\end{equation}
which is linear in $\bm{c}$. Stacking measurements gives the least-squares estimate $\bm{c}^0$ in the body frame. With $\bm{c}^0$ and $\tau_{\mathcal{I}}^0$ fixed, we solve
\begin{equation}
{}^{\mathcal{B}}\bm{R}_{\mathcal{I}}^0 = 
\arg\min_{\bm{R} \in \mathrm{SO}(3)}
\sum_{k} \left\|
\bm{z}^{\omega}_{\mathcal{I},k}-
\bm{R}^{\top}\bm{\omega}_{\mathcal{B},k}
\right\|^2,
\label{eq:gyro_wahba}
\end{equation}
using Kabsch alignment. The IMU-frame bias is recovered as $\bm{b}_g^0=({}^{\mathcal{B}}\bm{R}_{\mathcal{I}}^0)^{\top}\bm{c}^0$. These estimates initialize a bounded robust fit of the gyroscope residual
\begin{equation}
\bm{r}^{g}_{\mathcal{I},k}=
\bm{z}^{\omega}_{\mathcal{I},k}
-\;{}^{\mathcal I}\bm{\omega}_{\mathcal I,k}
-\bm{b}_g,
\label{eq:gyro_residual}
\end{equation}
over rotation, time offset, and gyroscope bias; the result is held fixed in the accelerometer stage.

\subsubsection{Accelerometer Initialization}

Left-multiplying \eqref{eq:imu_observation} by $\bm{R}_{\mathcal B\mathcal I}$ gives the body-frame measured specific force
\begin{equation}
\bm{a}^{\mathcal B}_{m,k}={}^{\mathcal B}\bm{R}_{\mathcal I}\bm{z}^{a}_{\mathcal I,k}
=s_a\left(\bm{u}_{r,k}+\bm{L}_k\bm{p}_{\mathcal I}-\bm{g}_{\mathcal B,k}\right)+\bm{b}_a^{\mathcal B}.
\label{eq:accel_body}
\end{equation}
Direct use of \eqref{eq:accel_body} requires the evolution of body-frame gravity and therefore global attitude integration, which accumulates drift and couples gravity with the accelerometer bias. For any body-frame vector $\bm{q}$, define the inertial time derivative resolved in the body frame,
\begin{equation}
\operatorname{D}\bm{q}=\dot{\bm{q}}+\bm{\omega}_{\mathcal B}\times\bm{q}.
\label{eq:D_operator}
\end{equation}
Here $\bm{a}^{\mathcal B}_{m,k}$ is the measured specific force. Rearranging \eqref{eq:accel_body} gives
\begin{equation}
\bm{g}_{\mathcal B,k}=
\bm{u}_{r,k}+\bm{L}_k\bm{p}_{\mathcal I}
-\frac{\bm{a}^{\mathcal B}_{m,k}-\bm{b}_a^{\mathcal B}}{s_a}.
\label{eq:gravity_isolated}
\end{equation}
Since gravity is fixed in the inertial frame, $\operatorname{D}\bm{g}_{\mathcal B}=\bm{0}_{3\times1}$. Applying \eqref{eq:D_operator} to \eqref{eq:gravity_isolated}, and noting that $\operatorname{D}\bm{b}_a^{\mathcal B}=[\bm{\omega}_{\mathcal B}]_\times\bm{b}_a^{\mathcal B}$ and $\operatorname{D}(\bm{L}_k\bm{p}_{\mathcal I})=(\dot{\bm{L}}_k+[\bm{\omega}_{\mathcal B}]_\times\bm{L}_k)\bm{p}_{\mathcal I}$, yields
\begin{equation}
\bm{0}_{3\times1}=
\operatorname{D}\bm{u}_{r}
+\left(\dot{\bm{L}}+\left[\bm{\omega}_{\mathcal B}\right]_\times\bm{L}\right)\bm{p}_{\mathcal I}
-\frac{\operatorname{D}\bm{a}^{\mathcal B}_{m}
-[\bm{\omega}_{\mathcal B}]_\times\bm{b}_a^{\mathcal B}}{s_a}.
\label{eq:gravity_derivative}
\end{equation}
Multiplying \eqref{eq:gravity_derivative} by $s_a$ gives the gravity-free relation
\begin{equation}
s_a\operatorname{D}\bm{u}_{r}
+s_a\left(\dot{\bm{L}}+\left[\bm{\omega}_{\mathcal B}\right]_\times\bm{L}\right)\bm{p}_{\mathcal I}
-\operatorname{D}\bm{a}^{\mathcal B}_{m}
=-\left[\bm{\omega}_{\mathcal B}\right]_\times\bm{b}_a^{\mathcal B}.
\label{eq:gravity_free_relation}
\end{equation}

\begin{figure}[t]
    \centering
\includegraphics[width=1.0\linewidth]{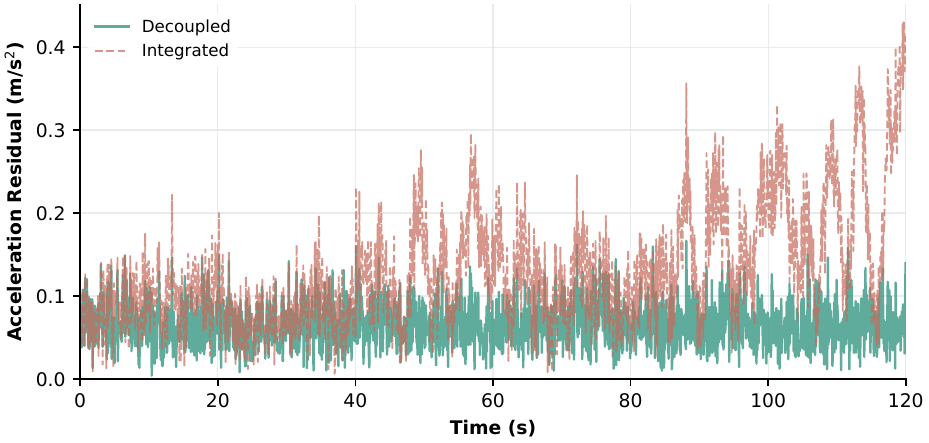}
    \caption{Simulated acceleration residuals for the decoupled and integrated gravity formulations.}
    \label{fig:gravity_simulation}
    \vspace{-0.6 cm}
\end{figure}

Integrating \eqref{eq:gravity_free_relation} over a sliding window $[t_i,t_j]$ absorbs the derivative into boundary terms. For a body-frame vector $\bm q$, define
\begin{equation}
\operatorname{W}_{ij}(\bm{q})=
\int_{t_i}^{t_j}\operatorname{D}\bm{q}\,dt
=\bm{q}_j-\bm{q}_i+\int_{t_i}^{t_j}\left[\bm{\omega}_{\mathcal B}\right]_\times\bm{q}\,dt.
\label{eq:W_operator}
\end{equation}
Let $\bm{y}_{m,ij}=\operatorname{W}_{ij}(\bm{a}^{\mathcal B}_{m})$ and $\bm{y}_{r,ij}=\operatorname{W}_{ij}(\bm{u}_r)$, and define the window matrices
\begin{equation}
\begin{aligned}
\bm{M}_{ij}&=\bm{L}_j-\bm{L}_i+
\int_{t_i}^{t_j}\left[\bm{\omega}_{\mathcal B}\right]_\times\bm{L}\,dt,\\
\bm{N}_{ij}&=-\int_{t_i}^{t_j}\left[\bm{\omega}_{\mathcal B}\right]_\times dt,
\end{aligned}
\label{eq:accel_window_matrices}
\end{equation}
Since $\operatorname{W}_{ij}(\bm{L}\bm{p}_{\mathcal I})=\bm{M}_{ij}\bm{p}_{\mathcal I}$, applying \eqref{eq:W_operator} to \eqref{eq:gravity_free_relation} gives
\begin{equation}
s_a\bm{y}_{r,ij}
+s_a\bm{M}_{ij}\bm{p}_{\mathcal I}
-\bm{y}_{m,ij}
=\bm{N}_{ij}\bm{b}_a^{\mathcal B}.
\label{eq:accel_integrated}
\end{equation}
Rearranging \eqref{eq:accel_integrated} yields
\begin{equation}
\bm{y}_{m,ij}=
s_a\bm{M}_{ij}\bm{p}_{\mathcal I}
-\bm{N}_{ij}\bm{b}_a^{\mathcal B}
+s_a\bm{y}_{r,ij}.
\label{eq:accel_window}
\end{equation}
To preserve linearity, let $\tilde{\bm{p}}_{\mathcal I}=s_a\bm{p}_{\mathcal I}$. Then
\begin{equation}
\bm{y}_{m,ij}=
\bm{M}_{ij}\tilde{\bm{p}}_{\mathcal I}
-\bm{N}_{ij}\bm{b}_a^{\mathcal B}
+s_a\bm{y}_{r,ij},
\label{eq:accel_lumped}
\end{equation}
and $\bm{p}_{\mathcal I}=\tilde{\bm{p}}_{\mathcal I}/s_a$. To retain physical consistency, we reconstruct gravity from \eqref{eq:gravity_isolated} using the estimated parameters,
\begin{equation}
\hat{\bm{g}}_{\mathcal B,k}=
\bm{u}_{r,k}+\bm{L}_k\bm{p}_{\mathcal I}
-\frac{\bm{a}^{\mathcal B}_{m,k}-\bm{b}_a^{\mathcal B}}{s_a}
\label{eq:gravity_reconstruction}
\end{equation}
and softly constrain $\|\hat{\bm{g}}_{\mathcal B,k}\|$ to $g_0=9.81\,\mathrm{m/s^2}$. The initialization solves the corresponding weighted least-squares problem; a rank-seven design matrix is required to resolve all parameters. The resulting values initialize the refinement stage below.

\subsubsection{Refinement}
\label{sec:imu_joint}

With rotation and time offset fixed at their gyroscope-stage estimates, we refine the lumped translation $\tilde{\bm{p}}_{\mathcal I}$, body-frame bias $\bm{b}_a^{\mathcal B}$, and scale $s_a$:
\begin{equation}
\begin{aligned}
\min_{\tilde{\bm p}_{\mathcal I},\bm b_a^{\mathcal B},s_a}\quad
&\sum_{i,j}\rho\!\left(w_{ij}\left\|\bm r^{a}_{ij}\right\|^2\right)
+\rho\!\left(\frac{\left\|\bm b_a^{\mathcal B}\right\|^2}{\sigma_b^2}\right)\\
&\qquad+\rho\!\left((s_a-1)^2\right),\\
\bm r^{a}_{ij}={}&\bm y_{m,ij}-\bm M_{ij}\tilde{\bm p}_{\mathcal I}
+\bm N_{ij}\bm b_a^{\mathcal B}-s_a\bm y_{r,ij},
\end{aligned}
\label{eq:imu_joint}
\end{equation}
where $\sigma_b$ is the bias prior scale and $w_{ij}$ is an angular-speed-based window weight. The implementation uses bounded trust-region least squares with a soft-$L_1$ loss and a weak preference for $s_a=1$. The physical lever arm is recovered as $\bm{p}_{\mathcal I}=\tilde{\bm{p}}_{\mathcal I}/s_a$. Gravity may be reconstructed for diagnostics but is not refined.

\subsection{Reference-Conditioned Block Refinement}
\label{sec:joint_calibration}

The modular calibrations in Sections~\ref{sec:radar_calibration} to \ref{sec:imu_calibration} use the same reference twist. Their estimates initialize a reference-conditioned refinement over $\bm{\theta}$. No direct sensor-to-sensor factor is present, so the objective factors sensor-wise exactly as in \eqref{eq:conditional_posterior}. Each block evaluates the reference at its aligned timestamps and is solved independently with bounded trust-region least squares.

In the gyroscope-only configuration, the IMU block uses $\bm{r}^{g}_{\mathcal I,k}$ from \eqref{eq:gyro_residual}; its translation, accelerometer bias, and scale remain at their modular-stage estimates. The accelerometer-enabled configuration, used only in the initialization ablation in Table~\ref{tab:optimization_ablation}, additionally includes the window residual in \eqref{eq:imu_joint}, with the residual re-evaluated at the current IMU rotation and time offset. No gravity state is optimized. One sensor block cannot resolve another block's unobservable translation component.

\section{Experiments}
\label{sec:experiments}
\subsection{Experimental Setup}
The platform is a rigidly mounted aerial sensing rig with two 4D mmWave radars, one event camera, and two IMUs. A LiDAR-inertial odometry system based on FAST-LIO~\cite{xu2021fastlio} provides the reference body twist only; no LiDAR points, planes, maps, or geometric registration enter the calibration residuals. The reference is therefore an external conditioning input, and all estimates are interpreted conditionally on its synchronization and accuracy. Its errors are common across sensor blocks rather than independent per-modality calibration errors. If the reference uses one of the calibrated IMUs, that sensor block is not independent of the reference; the independence assumption in Proposition~1 is therefore interpreted per sensor block.

\begin{figure*}[!t]
    \centering
    \begin{minipage}[t]{0.32\textwidth}
        \centering
        \includegraphics[width=\linewidth]{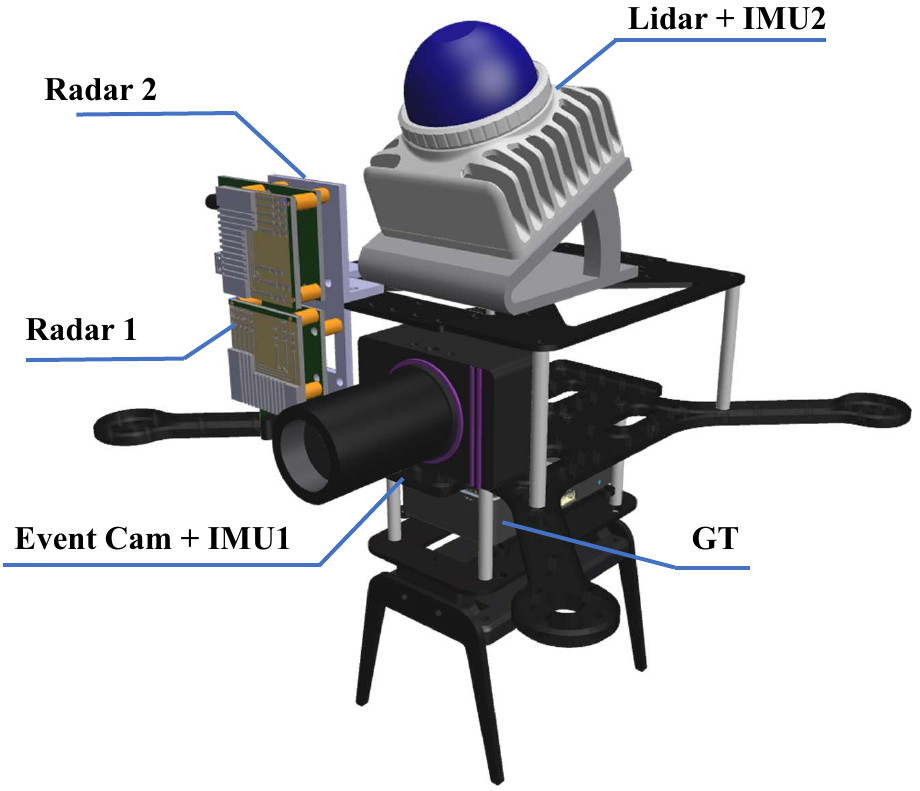}\\[-2pt]
        {\scriptsize (a)}
    \end{minipage}\hfill
    \begin{minipage}[t]{0.32\textwidth}
        \centering
        \includegraphics[width=\linewidth]{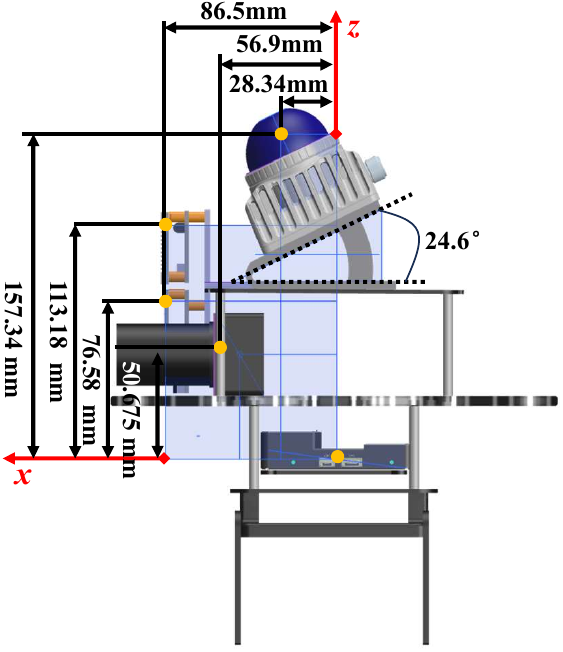}\\[-2pt]
        {\scriptsize (b)}
    \end{minipage}\hfill
    \begin{minipage}[t]{0.32\textwidth}
        \centering
        \includegraphics[width=\linewidth]{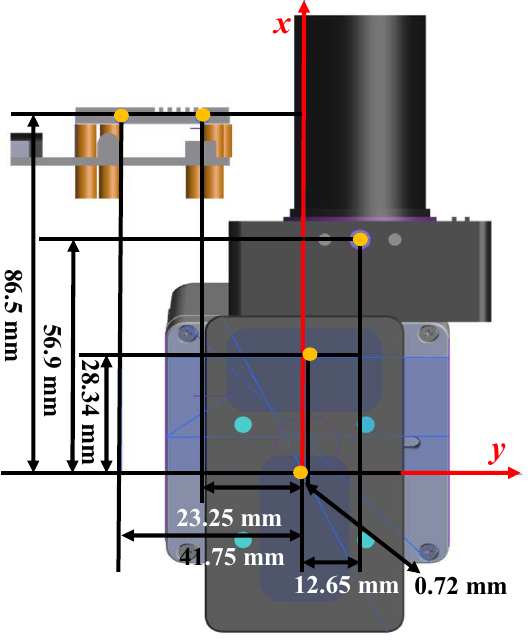}\\[-2pt]
        {\scriptsize (c)}
    \end{minipage}
    \caption{Experimental platform and coordinate definitions. (a) Sensor assembly. (b) X--Z and (c) X--Y layouts; dimensions in mm. The LiDAR is used only to provide the reference twist.}
    \label{fig:exp_platform}
\end{figure*}

The rig is hand-carried through full six-degree-of-freedom motion because the lever-arm term in \eqref{eq:lever} requires angular excitation. Fig.~\ref{fig:exp_platform} shows the sensor layout and coordinate definitions.

\begin{figure}[htbp] 
    \centering
    
    \begin{minipage}[c]{0.49\linewidth}
        \centering
        \vspace{-0.2cm}
        \begin{subfigure}{\linewidth}
            \centering
            \includegraphics[width=\linewidth]{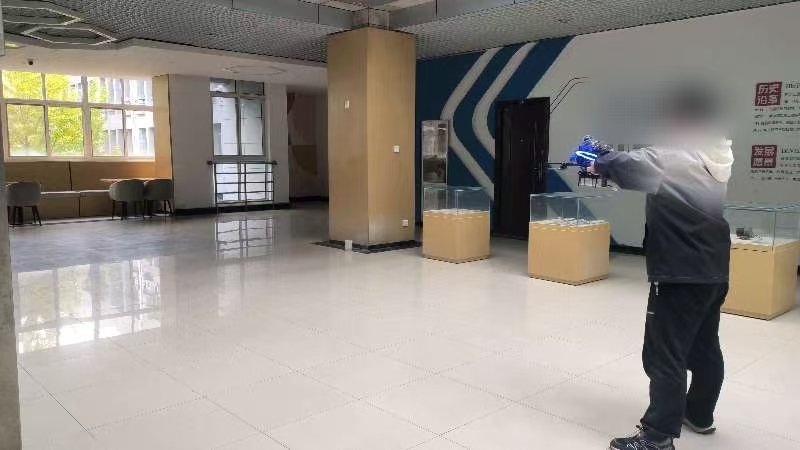} 
            \caption{Indoor scene}
            \label{fig:scene_indoor}
        \end{subfigure}
        
        
        \begin{subfigure}{\linewidth}
            \centering
            \includegraphics[width=\linewidth]{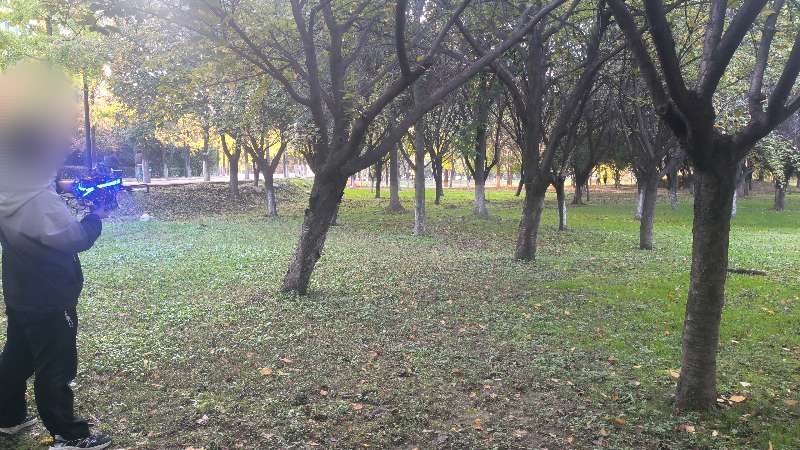} 
            \caption{Forest scene}
            \label{fig:scene_forest}
        \end{subfigure}
    \end{minipage}
    \hfill 
    \begin{minipage}[c]{0.49\linewidth}
        \centering
        \begin{subfigure}{\linewidth}
            \centering
            \includegraphics[width=\linewidth]{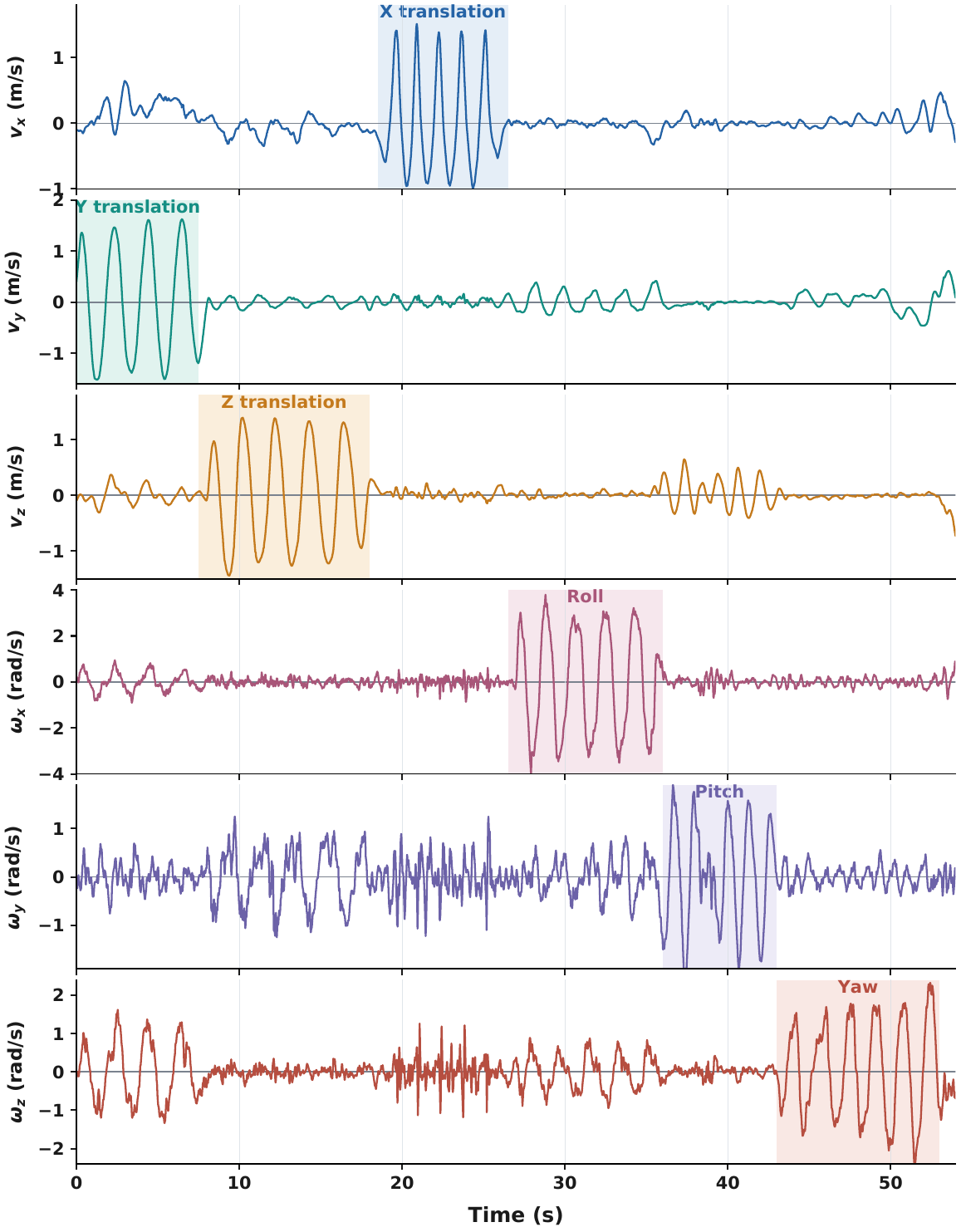} 
            \caption{Proposed axis-separated omnidirectional excitation}
            \label{fig:seq3_scene_twist}
        \end{subfigure}
    \end{minipage}
    
\caption{Data collection environments and motion. (a) Indoor scene; (b) forest scene; (c) reference linear and angular velocities for the axis-separated motion in Sequence~3.}
    \label{fig:data_collection_overview}
\end{figure}

Four hand-carried recordings are used. Seq.~1 and Seq.~2 are indoor translation- and rotation-dominant conditioning cases; they isolate the two single-type excitation limits rather than serving as complete calibration profiles. Seq.~3 is the proposed axis-separated omnidirectional excitation, with one rotation and one translation phase per axis. Seq.~4 repeats this protocol in a forest scene. The axis-separated motions avoid isotropic or overly compound trajectories that leave velocity-domain relative motion weakly identifiable. Scene, duration, and velocity percentiles are summarized in Table~\ref{tab:final_sequences}; the Seq.~3 profile is shown in Fig.~\ref{fig:seq3_scene_twist}.

\begin{table}[t]
\centering
\caption{Calibration recordings.}
\label{tab:final_sequences}
\scriptsize
\setlength{\tabcolsep}{4pt}
\begin{tabular}{rllrrr}
\toprule
Seq. & Scene & Motion & Dur. (s) & $v_{95}$ (m/s) & $\omega_{95}$ (rad/s) \\
\midrule
1 & indoor & translation-dominant & 22.48 & 1.415 & 1.239 \\
2 & indoor & rotation-dominant    & 20.98 & 0.368 & 2.541 \\
3 & indoor & axis-separated   & 53.98 & 1.381 & 2.810 \\
4 & forest & axis-separated   & 57.48 & 1.586 & 2.522 \\
\bottomrule
\end{tabular}
\par\vspace{2pt}\begin{minipage}{\linewidth}\scriptsize\raggedright \textit{Note:} Sequences 1--4 correspond to 10-29/1, 10-29/2, 10-29/6, and 10-30/1. Duration spans the reference-twist timestamps; percentiles use speed norms.\end{minipage}
\end{table}

The two IMUs are indexed as \(\mathcal{I}_1\) (Livox) and \(\mathcal{I}_2\) (DVS), with acceleration and angular-rate modalities \(\mathcal{I}_{a,i}\) and \(\mathcal{I}_{\omega,i}\). For fair comparison, all calibration states are reported relative to the DVS-IMU frame \(\mathcal{I}_2\), since the baselines conventionally adopt an IMU frame. The sensor-to-reference extrinsics are transformed as \(\bm{x}_{\mathcal{I}_2}=\bm{R}_{\mathcal{I}_2,\mathcal{S}}\bm{x}_{\mathcal{S}}+\bm{p}_{\mathcal{I}_2,\mathcal{S}}\), where \(\bm{p}_{\mathcal{I}_2,\mathcal{S}}\) runs from the DVS-IMU origin to the sensor origin. Rotations are reported as \(\bm{r}=\log(\bm{R}_{\mathcal{I}_2,\mathcal{S}})^\vee\) in degrees, translations in centimeters, and relative time offsets as \(t_{\mathcal{I}_2}=t_{\mathcal{S}}-\tau_{\mathcal{S}/\mathcal{I}_2}\) with \(\tau_{\mathcal{S}/\mathcal{I}_2}=\tau_{\mathcal{S}/\mathrm{ref}}-\tau_{\mathcal{I}_2/\mathrm{ref}}\).

The recordings use independent clocks, so absolute offsets vary with the recording epoch and are not physical delay differences across sessions. The DVS-IMU identity is a reporting convention. CAD is an assembly reference rather than metrology, has no measured offset, and its point-cloud axis mapping requires independent verification. The reported \(1\sigma\) values are local covariance estimates rather than repeated-run statistics.

The main calibration tables report $\bm{\theta}_{\mathcal S}^{\mathrm{st}}$, while the augmented parameters $\bm{\beta}_{\mathcal R}$ and $\bm{\beta}_{\mathcal I}$ are isolated in the bias/scale ablation of Table~\ref{tab:sensor_parameter_ablation}. Their definitions are identical to \eqref{eq:augmented_params}: radar bias and scale are $\bm{b}_{\mathcal R}^{\mathcal B}$ and $s_{\mathcal R}$, and IMU gyroscope bias, accelerometer bias, and accelerometer scale are $\bm{b}_g$, $\bm{b}_a^{\mathcal B}$, and $s_a$.

We report RMSE of frozen-calibration residuals on shared observations: radar velocity, event normal flow, IMU gyroscope, and IMU acceleration preintegral. These residuals measure constraint consistency, unlike local STD, which describes precision. Shared nuisance parameters are held fixed for all methods without refitting; CAD\(^*\) uses Ours' timing and is not ranked.

\subsection{Calibration Results}

\noindent\textit{Baselines and protocol.} Radar results are compared with RIs-Calib~\cite{riscalib2024} and iKalibr~\cite{chen2025ikalibr}; event results with RTEI-Calib~\cite{rtei2024} and EvMultiCalib~\cite{mai2025eventcalib}; and IMU results with MI-Calib~\cite{chen2024mi} and iKalibr~\cite{chen2025ikalibr}. Each method receives the same recordings and, where accepted, the same synchronized reference twist; it is run with its native state model, preprocessing, active interval, and default initialization. A state not produced by a baseline is marked unavailable rather than supplied from our estimate. All states are expressed relative to the DVS-IMU frame \(\mathcal I_2\) as described above. We report the estimate and local $1\sigma$ from each solver, and use the CAD assembly only as a Sequence~3 sanity reference.

Each modality is reported in the same order: the estimate and local covariance, the corresponding alignment and residual diagnostics, and the measurement-model explanation for the observed uncertainty. This keeps numerical variability separate from its physical cause.

\subsubsection{Radar \(\mathcal R_1\) Calibration}

\begin{figure}[t]
    \centering
    \includegraphics[width=\linewidth]{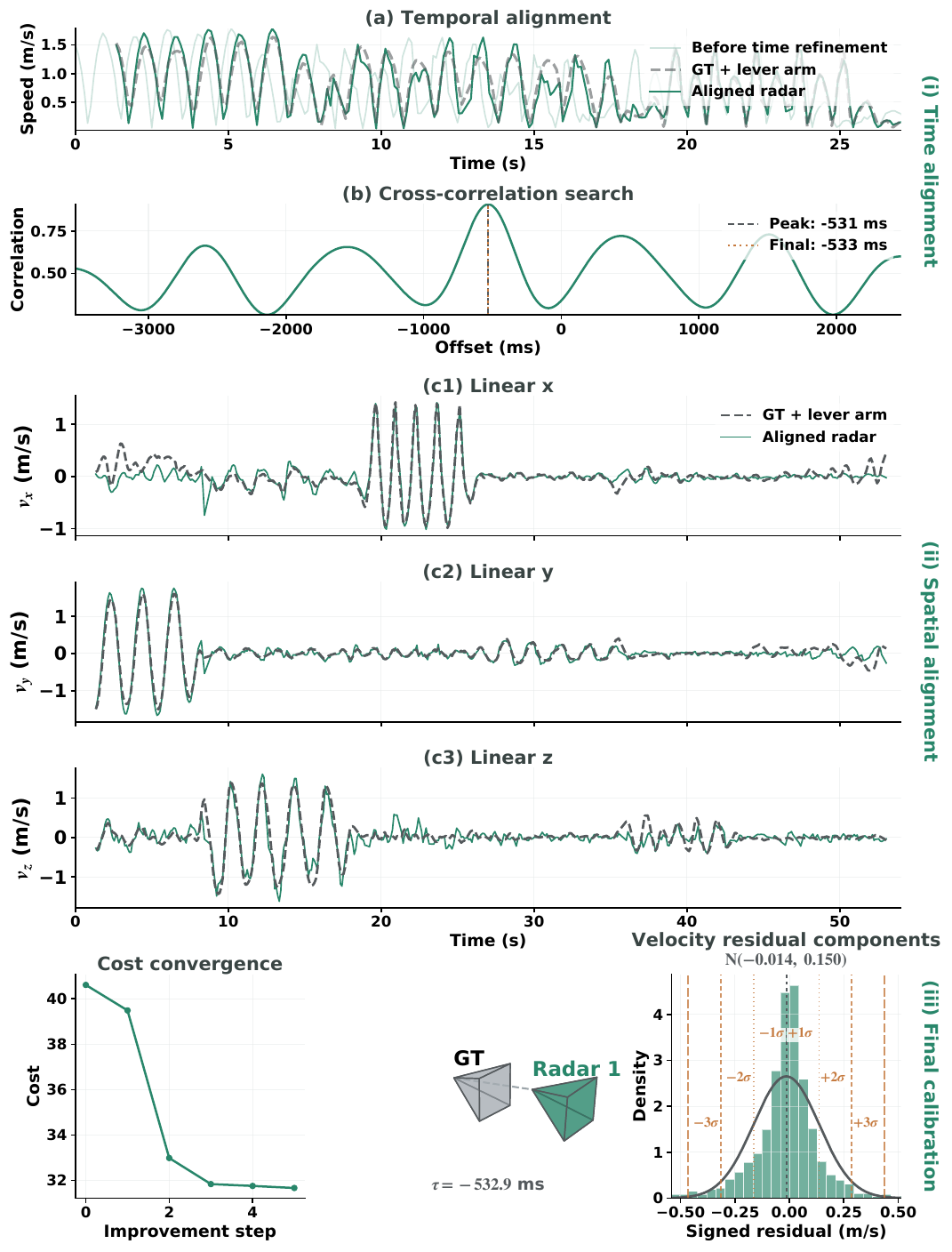}
    \caption{Radar 1 calibration diagnostics for Sequence~3. (a) Speed alignment; (b) offset correlation; (c1)--(c3) velocity components. The bottom row shows optimization cost, sensor geometry, and velocity residuals. The offset is relative to the reference twist frame.}
    \label{fig:radar1_pipeline}
\end{figure}

Table~\ref{tab:modality_radar_main} and Fig.~\ref{fig:radar1_pipeline} report the \(\mathcal R_1\) estimate and its Sequence~3 diagnostics.

\begin{table}[!t]
\centering
\caption{Calibration results for radar \(\mathcal R_1\).}
\label{tab:modality_radar_main}
\scriptsize
\setlength{\tabcolsep}{1.5pt}
\renewcommand{\arraystretch}{0.92}
\begin{tabular*}{\columnwidth}{@{\extracolsep{\fill}}llr@{}lr@{}lr@{}l}
\toprule
Seq. & State & \multicolumn{2}{c}{Ours} & \multicolumn{2}{c}{RIs-Calib~\cite{riscalib2024}} & \multicolumn{2}{c}{iKalibr~\cite{chen2025ikalibr}} \\
\midrule
\multirow{7}{*}{1} & $p_{x}$ & $5.44$ & ${}\,\pm\,\mathbf{6.08}$ & $6.33$ & ${}\,\pm\,7.55$ & $8.54$ & ${}\,\pm\,9.86$ \\
 & $p_{y}$ & $12.21$ & ${}\,\pm\,\mathbf{6.39}$ & $16.55$ & ${}\,\pm\,7.15$ & $16.27$ & ${}\,\pm\,7.38$ \\
 & $p_{z}$ & $-3.61$ & ${}\,\pm\,4.94$ & $-1.74$ & ${}\,\pm\,\mathbf{2.33}$ & $-2.76$ & ${}\,\pm\,3.56$ \\
 & $r_{x}$ & $67.75$ & ${}\,\pm\,\mathbf{1.19}$ & $73.17$ & ${}\,\pm\,1.77$ & $66.96$ & ${}\,\pm\,1.34$ \\
 & $r_{y}$ & $-65.83$ & ${}\,\pm\,\mathbf{1.27}$ & $-60.11$ & ${}\,\pm\,2.79$ & $-89.66$ & ${}\,\pm\,9.98$ \\
 & $r_{z}$ & $72.27$ & ${}\,\pm\,\mathbf{1.37}$ & $46.60$ & ${}\,\pm\,11.79$ & $46.50$ & ${}\,\pm\,11.94$ \\
 & $\tau$ & $287.13$ & ${}\,\pm\,\mathbf{3.61}$ & $280.76$ & ${}\,\pm\,4.92$ & $277.98$ & ${}\,\pm\,7.86$ \\
\midrule
\multirow{7}{*}{2} & $p_{x}$ & $1.05$ & ${}\,\pm\,\mathbf{2.00}$ & $0.64$ & ${}\,\pm\,7.86$ & $0.85$ & ${}\,\pm\,5.10$ \\
 & $p_{y}$ & $-5.16$ & ${}\,\pm\,\mathbf{1.51}$ & $7.52$ & ${}\,\pm\,3.77$ & $-2.19$ & ${}\,\pm\,4.63$ \\
 & $p_{z}$ & $-3.68$ & ${}\,\pm\,\mathbf{2.88}$ & $-7.57$ & ${}\,\pm\,3.71$ & $-8.03$ & ${}\,\pm\,5.27$ \\
 & $r_{x}$ & $73.11$ & ${}\,\pm\,9.85$ & $71.53$ & ${}\,\pm\,6.90$ & $62.09$ & ${}\,\pm\,\mathbf{6.15}$ \\
 & $r_{y}$ & $-83.79$ & ${}\,\pm\,9.59$ & $-72.74$ & ${}\,\pm\,\mathbf{4.69}$ & $-78.20$ & ${}\,\pm\,6.23$ \\
 & $r_{z}$ & $64.33$ & ${}\,\pm\,\mathbf{7.21}$ & $83.64$ & ${}\,\pm\,8.44$ & $85.21$ & ${}\,\pm\,9.30$ \\
 & $\tau$ & $281.54$ & \multicolumn{1}{c}{--} & $280.73$ & ${}\,\pm\,2.21$ & $286.35$ & ${}\,\pm\,6.09$  \\
\midrule
\multirow{7}{*}{3} & $p_{x}$ & $3.89$ & ${}\,\pm\,\mathbf{1.75}$ & $4.68$ & ${}\,\pm\,2.87$ & $7.11$ & ${}\,\pm\,3.61$ \\
 & $p_{y}$ & $-1.91$ & ${}\,\pm\,\mathbf{1.62}$ & $6.27$ & ${}\,\pm\,7.28$ & $7.61$ & ${}\,\pm\,8.20$ \\
 & $p_{z}$ & $3.99$ & ${}\,\pm\,\mathbf{2.66}$ & $-3.79$ & ${}\,\pm\,9.96$ & $6.09$ & ${}\,\pm\,10.85$ \\
 & $r_{x}$ & $69.46$ & ${}\,\pm\,\mathbf{0.83}$ & $63.35$ & ${}\,\pm\,7.36$ & $69.87$ & ${}\,\pm\,1.10$ \\
 & $r_{y}$ & $-69.46$ & ${}\,\pm\,\mathbf{0.90}$ & $-63.12$ & ${}\,\pm\,6.88$ & $-73.88$ & ${}\,\pm\,12.29$ \\
 & $r_{z}$ & $68.74$ & ${}\,\pm\,\mathbf{0.86}$ & $71.50$ & ${}\,\pm\,2.96$ & $64.61$ & ${}\,\pm\,4.38$ \\
 & $\tau$ & $235.48$ & ${}\,\pm\,\mathbf{2.39}$ & $252.08$ & ${}\,\pm\,15.62$ & $236.05$ & ${}\,\pm\,2.60$ \\
\midrule
\multirow{7}{*}{4} & $p_{x}$ & $4.20$ & ${}\,\pm\,\mathbf{1.87}$ & $6.44$ & ${}\,\pm\,5.19$ & $6.41$ & ${}\,\pm\,4.42$ \\
 & $p_{y}$ & $-4.49$ & ${}\,\pm\,\mathbf{1.80}$ & $-6.33$ & ${}\,\pm\,4.73$ & $-5.65$ & ${}\,\pm\,2.97$ \\
 & $p_{z}$ & $4.60$ & ${}\,\pm\,\mathbf{2.39}$ & $5.58$ & ${}\,\pm\,3.64$ & $4.43$ & ${}\,\pm\,2.52$ \\
 & $r_{x}$ & $69.70$ & ${}\,\pm\,\mathbf{0.68}$ & $52.02$ & ${}\,\pm\,12.74$ & $66.59$ & ${}\,\pm\,1.44$ \\
 & $r_{y}$ & $-72.33$ & ${}\,\pm\,\mathbf{0.73}$ & $-63.72$ & ${}\,\pm\,5.73$ & $-65.25$ & ${}\,\pm\,4.82$ \\
 & $r_{z}$ & $67.65$ & ${}\,\pm\,\mathbf{0.67}$ & $84.86$ & ${}\,\pm\,15.79$ & $63.15$ & ${}\,\pm\,4.34$ \\
 & $\tau$ & $288.58$ & ${}\,\pm\,\mathbf{1.99}$ & $276.02$ & ${}\,\pm\,10.59$ & $284.85$ & ${}\,\pm\,4.23$ \\
\bottomrule
\end{tabular*}
\vspace{2pt}
\par\begin{minipage}{\linewidth}\scriptsize\raggedright \textit{Note:} Units: $\bm{p}$ cm, $\bm{r}$ deg, $\tau$ ms. Entries are estimate $\pm$ local $1\sigma$; bold marks the smallest STD. The Seq.~2 offset is fixed at 281.54 ms and is not ranked. CAD (Seq.~3): $\bm{p}=(5.45,-2.59,2.96)$ cm and $\bm{r}=(69.28,-69.28,69.28)^\circ$.\end{minipage}
\end{table}

Radar \(\mathcal R_1\) maintains local time STDs of 1.99--3.61 ms for the ranked sequences, and its velocity envelope follows the reference after correlation compensation. The component traces in Fig.~\ref{fig:radar1_pipeline}(c1)--(c3) retain the lever-arm-induced phase relation rather than matching only the speed magnitude. In Sequence~3, the recovered lever arm is component-wise within 1.6 cm and 0.6$^\circ$ of the CAD assembly. The \(p_x\) STD decreases from 6.08 cm in translation-dominant Sequence~1 to 1.75 and 1.87 cm in Sequences~3 and~4, consistent with the lever-arm excitation condition in Section~\ref{subsec:schur_conditioning}. Absolute offsets differ across recordings because the reference and radar streams are acquired by two computers with different clock epochs and device latencies; they are not interpreted as repeatability. The Sequence~3 residual histogram is approximately zero-mean; this supports model consistency rather than unbiasedness of the extrinsics.

\subsubsection{Radar \(\mathcal R_2\) Calibration}

\begin{figure}[t]
    \centering
    \includegraphics[width=\linewidth]{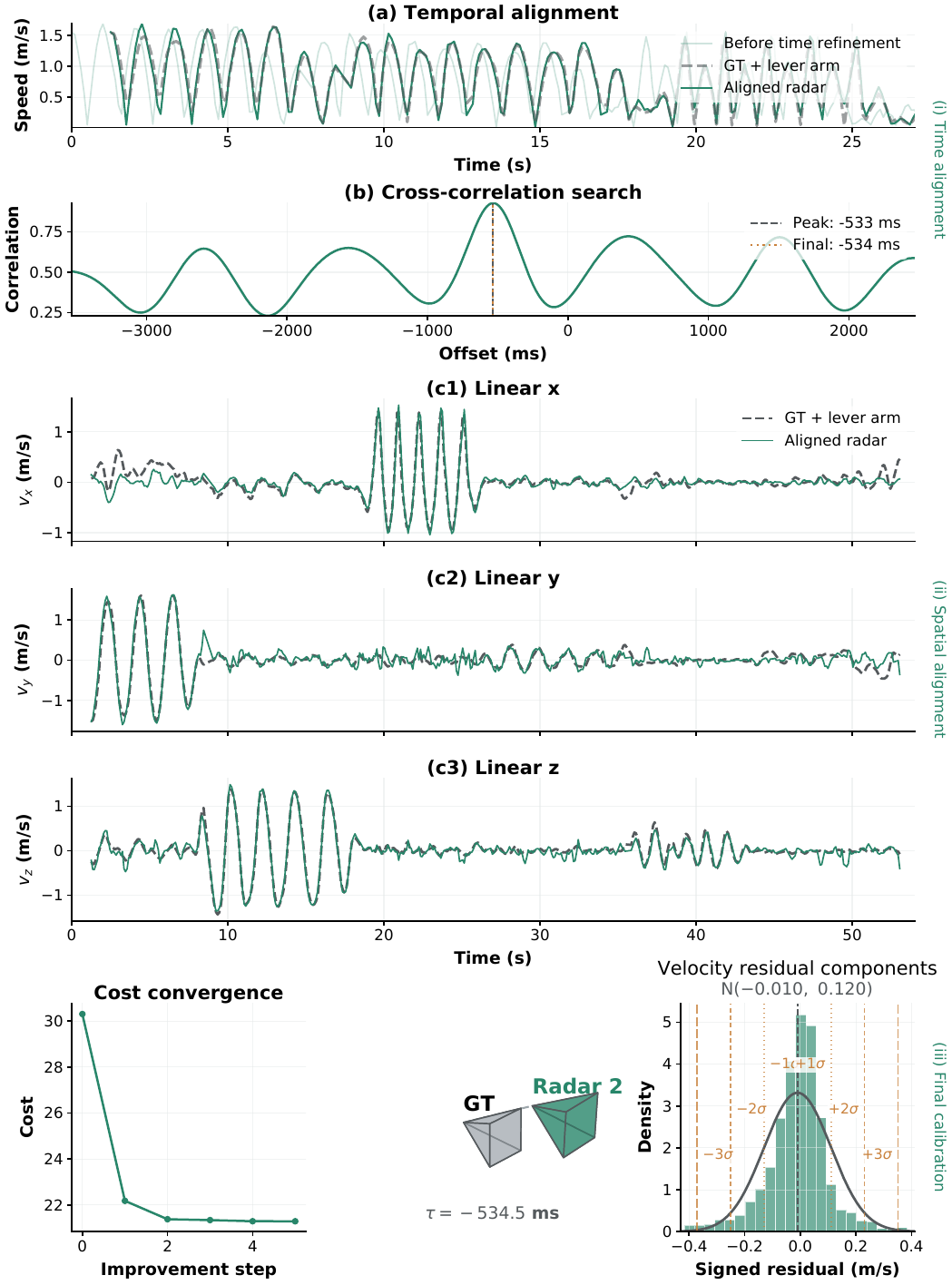}
    \caption{Radar 2 calibration diagnostics for Sequence~3. (a) Speed alignment; (b) offset correlation; (c1)--(c3) velocity components. The bottom row shows optimization cost, sensor geometry, and velocity residuals. The offset is relative to the reference twist frame.}
    \label{fig:radar2_pipeline}
\end{figure}

Table~\ref{tab:modality_radar_aux} and Fig.~\ref{fig:radar2_pipeline} report the corresponding \(\mathcal R_2\) estimate and diagnostics.

\begin{table}[!t]
\centering
\caption{Calibration results for radar \(\mathcal R_2\).}
\label{tab:modality_radar_aux}
\scriptsize
\setlength{\tabcolsep}{1.5pt}
\renewcommand{\arraystretch}{0.92}
\begin{tabular*}{\columnwidth}{@{\extracolsep{\fill}}llr@{}lr@{}lr@{}l}
\toprule
Seq. & State & \multicolumn{2}{c}{Ours} & \multicolumn{2}{c}{RIs-Calib~\cite{riscalib2024}} & \multicolumn{2}{c}{iKalibr~\cite{chen2025ikalibr}} \\
\midrule
\multirow{7}{*}{1} & $p_{x}$ & $-0.10$ & ${}\,\pm\,5.59$ & $1.00$ & ${}\,\pm\,\mathbf{3.31}$ & $1.87$ & ${}\,\pm\,4.81$ \\
 & $p_{y}$ & $-1.91$ & ${}\,\pm\,5.17$ & $1.65$ & ${}\,\pm\,\mathbf{4.93}$ & $6.46$ & ${}\,\pm\,5.44$ \\
 & $p_{z}$ & $-2.84$ & ${}\,\pm\,4.76$ & $-1.74$ & ${}\,\pm\,5.12$ & $-3.13$ & ${}\,\pm\,\mathbf{4.12}$ \\
 & $r_{x}$ & $-1.15$ & ${}\,\pm\,\mathbf{0.92}$ & $-3.21$ & ${}\,\pm\,5.27$ & $-7.71$ & ${}\,\pm\,6.61$ \\
 & $r_{y}$ & $-87.21$ & ${}\,\pm\,\mathbf{1.06}$ & $-79.83$ & ${}\,\pm\,3.60$ & $-93.50$ & ${}\,\pm\,1.14$ \\
 & $r_{z}$ & $-0.13$ & ${}\,\pm\,\mathbf{0.88}$ & $-3.95$ & ${}\,\pm\,2.28$ & $5.33$ & ${}\,\pm\,2.65$ \\
 & $\tau$ & $282.40$ & ${}\,\pm\,\mathbf{2.96}$ & $260.22$ & ${}\,\pm\,8.32$ & $265.24$ & ${}\,\pm\,4.48$ \\
\midrule
\multirow{7}{*}{2} & $p_{x}$ & $4.39$ & ${}\,\pm\,\mathbf{1.32}$ & $0.39$ & ${}\,\pm\,2.69$ & $10.51$ & ${}\,\pm\,4.91$ \\
 & $p_{y}$ & $-5.42$ & ${}\,\pm\,1.29$ & $1.33$ & ${}\,\pm\,2.37$ & $1.19$ & ${}\,\pm\,\mathbf{1.21}$ \\
 & $p_{z}$ & $1.83$ & ${}\,\pm\,\mathbf{1.84}$ & $7.59$ & ${}\,\pm\,2.85$ & $8.86$ & ${}\,\pm\,4.77$ \\
 & $r_{x}$ & $-2.22$ & ${}\,\pm\,2.37$ & $8.91$ & ${}\,\pm\,\mathbf{1.31}$ & $9.27$ & ${}\,\pm\,3.64$ \\
 & $r_{y}$ & $-81.79$ & ${}\,\pm\,2.69$ & $-94.34$ & ${}\,\pm\,\mathbf{2.18}$ & $-77.89$ & ${}\,\pm\,6.56$ \\
 & $r_{z}$ & $2.19$ & ${}\,\pm\,2.18$ & $-6.77$ & ${}\,\pm\,4.76$ & $1.26$ & ${}\,\pm\,\mathbf{1.89}$ \\
 & $\tau$ & $257.60$ & ${}\,\pm\,\mathbf{4.85}$ & $256.35$ & ${}\,\pm\,8.84$ & $274.92$ & ${}\,\pm\,6.87$ \\
\midrule
\multirow{7}{*}{3} & $p_{x}$ & $4.27$ & ${}\,\pm\,\mathbf{1.71}$ & $1.09$ & ${}\,\pm\,2.48$ & $3.80$ & ${}\,\pm\,2.03$ \\
 & $p_{y}$ & $-5.76$ & ${}\,\pm\,\mathbf{1.58}$ & $-10.69$ & ${}\,\pm\,3.28$ & $-10.61$ & ${}\,\pm\,2.74$ \\
 & $p_{z}$ & $3.63$ & ${}\,\pm\,\mathbf{2.56}$ & $-3.78$ & ${}\,\pm\,2.76$ & $-2.08$ & ${}\,\pm\,3.47$ \\
 & $r_{x}$ & $0.76$ & ${}\,\pm\,\mathbf{0.59}$ & $-6.79$ & ${}\,\pm\,4.07$ & $2.75$ & ${}\,\pm\,3.70$ \\
 & $r_{y}$ & $-87.93$ & ${}\,\pm\,\mathbf{0.68}$ & $-86.34$ & ${}\,\pm\,1.45$ & $-84.24$ & ${}\,\pm\,2.45$ \\
 & $r_{z}$ & $-1.63$ & ${}\,\mathbf{\pm\,0.59}$ & $2.11$ & ${}\,\pm\,0.81$ & $8.65$ & ${}\,\pm\,3.61$ \\
 & $\tau$ & $233.91$ & ${}\,\pm\,\mathbf{1.79}$ & $242.63$ & ${}\,\pm\,4.96$ & $241.35$ & ${}\,\pm\,3.77$ \\
\midrule
\multirow{7}{*}{4} & $p_{x}$ & $4.83$ & ${}\,\pm\,\mathbf{1.85}$ & $6.19$ & ${}\,\pm\,4.90$ & $7.49$ & ${}\,\pm\,4.63$ \\
 & $p_{y}$ & $-6.41$ & ${}\,\pm\,\mathbf{1.78}$ & $-8.55$ & ${}\,\pm\,5.06$ & $0.07$ & ${}\,\pm\,3.57$ \\
 & $p_{z}$ & $-2.08$ & ${}\,\pm\,\mathbf{2.36}$ & $-8.42$ & ${}\,\pm\,4.87$ & $-8.02$ & ${}\,\pm\,6.50$ \\
 & $r_{x}$ & $0.28$ & ${}\,\pm\,\mathbf{0.49}$ & $2.93$ & ${}\,\pm\,1.79$ & $-1.26$ & ${}\,\pm\,2.14$ \\
 & $r_{y}$ & $-87.88$ & ${}\,\pm\,\mathbf{0.56}$ & $-79.37$ & ${}\,\pm\,2.15$ & $-79.53$ & ${}\,\pm\,1.94$ \\
 & $r_{z}$ & $-1.29$ & ${}\,\pm\,\mathbf{0.47}$ & $-3.50$ & ${}\,\pm\,3.77$ & $-3.87$ & ${}\,\pm\,4.70$ \\
 & $\tau$ & $289.50$ & ${}\,\pm\,\mathbf{0.90}$ & $278.41$ & ${}\,\pm\,8.36$ & $296.01$ & ${}\,\pm\,7.07$ \\
\bottomrule
\end{tabular*}
\vspace{2pt}
\par\begin{minipage}{\linewidth}\scriptsize\raggedright \textit{Note:} Units and notation as in Table~\ref{tab:modality_radar_main}. Radar time STD is recomputed at the saved solution. CAD (Seq.~3): $\bm{p}=(3.60,-6.25,2.96)$ cm and $\bm{r}=(0,-90,0)^\circ$; -- is unavailable.\end{minipage}
\end{table}

Radar \(\mathcal R_2\) exhibits the same temporal behavior, with a 0.90 ms local time STD in Sequence~4. Its Sequence~3 lever arm is component-wise within 0.7 cm and 2.1$^\circ$ of its own CAD pose, and the \(p_x\) STD falls from 5.59 cm in Sequence~1 to 1.71 and 1.85 cm in Sequences~3 and~4. The numerical estimates for the two radars need not coincide because their frames and assembly poses differ; each is compared with its own CAD reference. The larger rotation discrepancy and the Sequence~1 translation spread are consistent with the differing mounting orientation, limited field of view, and weaker lever-arm excitation under the same recorded motion.

\subsubsection{Event Camera Calibration}

\begin{figure}[h]
    \centering
    \includegraphics[width=\linewidth]{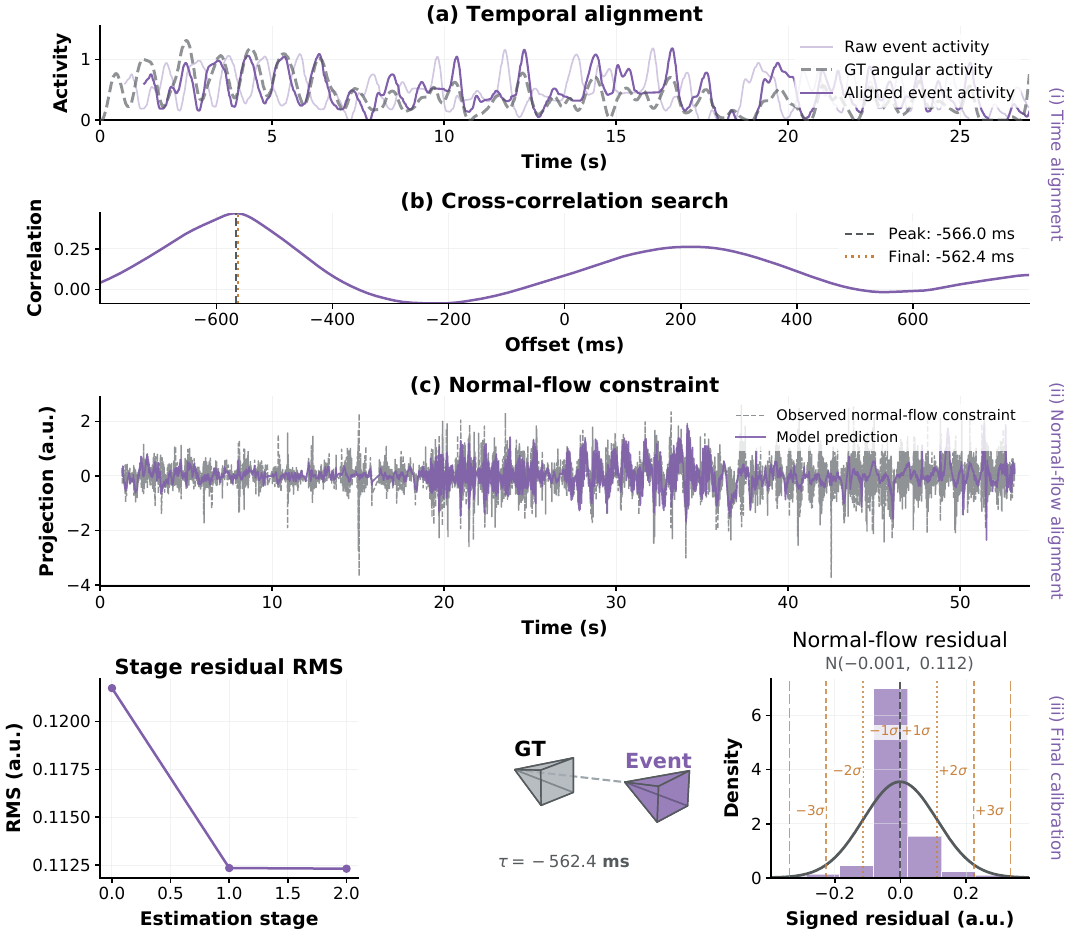}
    \caption{Event-camera calibration diagnostics for Sequence~3. (a) Activity alignment; (b) offset correlation; (c) normal-flow constraint. The bottom row shows stage residuals, sensor geometry, and residual distribution. The offset is relative to the reference twist frame.}
    \label{fig:event_pipeline}
\end{figure}

Table~\ref{tab:modality_event} and Fig.~\ref{fig:event_pipeline} report the event-camera results. Coupling normal flow with the reference twist recovers the full 6-DoF extrinsic, including the 3D lever arm that RTEI-Calib and EvMultiCalib do not provide, without scene reconstruction or frame-based pose estimation.

\begin{table}[!t]
\centering
\caption{Event camera calibration results.}
\label{tab:modality_event}
\scriptsize
\setlength{\tabcolsep}{1.5pt}
\renewcommand{\arraystretch}{0.92}
\begin{tabular*}{\columnwidth}{@{\extracolsep{\fill}}llr@{}lr@{}lr@{}l}
\toprule
Seq. & State & \multicolumn{2}{c}{Ours} & \multicolumn{2}{c}{RTEI-Calib~\cite{rtei2024}} & \multicolumn{2}{c}{EvMultiCalib~\cite{mai2025eventcalib}} \\
\midrule
\multirow{7}{*}{1} & $p_{x}$ & $22.35$ & ${}\,\pm\,\mathbf{4.94}$ & \multicolumn{2}{c}{--} & \multicolumn{2}{c}{--} \\
 & $p_{y}$ & $-2.28$ & ${}\,\pm\,\mathbf{4.59}$ & \multicolumn{2}{c}{--} & \multicolumn{2}{c}{--} \\
 & $p_{z}$ & $103.19$ & ${}\,\pm\,\mathbf{4.48}$ & \multicolumn{2}{c}{--} & \multicolumn{2}{c}{--} \\
 & $r_{x}$ & $35.66$ & ${}\,\pm\,\mathbf{0.50}$ & $11.46$ & ${}\,\pm\,0.73$ & $3.10$ & ${}\,\pm\,0.58$ \\
 & $r_{y}$ & $1.38$ & ${}\,\pm\,\mathbf{0.39}$ & $3.52$ & ${}\,\pm\,0.54$ & $3.05$ & ${}\,\pm\,0.71$ \\
 & $r_{z}$ & $4.50$ & ${}\,\pm\,\mathbf{0.36}$ & $2.10$ & ${}\,\pm\,0.38$ & $-1.67$ & ${}\,\pm\,0.44$ \\
 & $\tau$ & $198.75$ & ${}\,\pm\,\mathbf{0.43}$ & $207.08$ & ${}\,\pm\,0.79$ & $203.55$ & ${}\,\pm\,0.55$ \\
\midrule
\multirow{7}{*}{2} & $p_{x}$ & $4.19$ & ${}\,\pm\,\mathbf{1.10}$ & \multicolumn{2}{c}{--} & \multicolumn{2}{c}{--} \\
 & $p_{y}$ & $4.27$ & ${}\,\pm\,\mathbf{1.05}$ & \multicolumn{2}{c}{--} & \multicolumn{2}{c}{--} \\
 & $p_{z}$ & $5.86$ & ${}\,\pm\,\mathbf{1.62}$ & \multicolumn{2}{c}{--} & \multicolumn{2}{c}{--} \\
 & $r_{x}$ & $0.47$ & ${}\,\pm\,\mathbf{0.37}$ & $3.91$ & ${}\,\pm\,0.85$ & $3.40$ & ${}\,\pm\,0.53$ \\
 & $r_{y}$ & $-95.44$ & ${}\,\pm\,\mathbf{0.19}$ & $1.17$ & ${}\,\pm\,0.58$ & $-1.80$ & ${}\,\pm\,0.53$ \\
 & $r_{z}$ & $11.41$ & ${}\,\pm\,\mathbf{0.31}$ & $-0.85$ & ${}\,\pm\,0.68$ & $-0.06$ & ${}\,\pm\,0.66$ \\
 & $\tau$ & $205.52$ & ${}\,\pm\,\mathbf{0.71}$ & $207.07$ & ${}\,\pm\,1.25$ & $205.54$ & ${}\,\pm\,0.80$ \\
\midrule
\multirow{7}{*}{3} & $p_{x}$ & $0.16$ & ${}\,\pm\,\mathbf{1.66}$ & \multicolumn{2}{c}{--} & \multicolumn{2}{c}{--} \\
 & $p_{y}$ & $0.76$ & ${}\,\pm\,\mathbf{1.52}$ & \multicolumn{2}{c}{--} & \multicolumn{2}{c}{--} \\
 & $p_{z}$ & $1.41$ & ${}\,\pm\,\mathbf{2.48}$ & \multicolumn{2}{c}{--} & \multicolumn{2}{c}{--} \\
 & $r_{x}$ & $3.14$ & ${}\,\pm\,\mathbf{0.41}$ & $3.15$ & ${}\,\pm\,0.59$ & $2.29$ & ${}\,\pm\,0.77$ \\
 & $r_{y}$ & $1.84$ & ${}\,\pm\,\mathbf{0.37}$ & $1.94$ & ${}\,\pm\,0.84$ & $3.51$ & ${}\,\pm\,0.95$ \\
 & $r_{z}$ & $-0.30$ & ${}\,\pm\,\mathbf{0.38}$ & $-0.10$ & ${}\,\pm\,0.90$ & $-1.24$ & ${}\,\pm\,0.80$ \\
 & $\tau$ & $206.04$ & ${}\,\mathbf{\pm\,1.04}$ & $205.58$ & ${}\,\pm\,1.53$ & $206.55$ & ${}\,\pm\,1.67$ \\
\midrule
\multirow{7}{*}{4} & $p_{x}$ & $1.26$ & ${}\,\pm\,\mathbf{1.86}$ & \multicolumn{2}{c}{--} & \multicolumn{2}{c}{--} \\
 & $p_{y}$ & $1.19$ & ${}\,\pm\,\mathbf{1.78}$ & \multicolumn{2}{c}{--} & \multicolumn{2}{c}{--} \\
 & $p_{z}$ & $-4.38$ & ${}\,\pm\,\mathbf{2.39}$ & \multicolumn{2}{c}{--} & \multicolumn{2}{c}{--} \\
 & $r_{x}$ & $7.77$ & ${}\,\mathbf{\pm\,0.89}$ & $13.98$ & ${}\,\pm\,2.57$ & $15.24$ & ${}\,\pm\,2.87$ \\
 & $r_{y}$ & $-1.77$ & ${}\,\mathbf{\pm\,0.93}$ & $-3.93$ & ${}\,\pm\,2.72$ & $-4.87$ & ${}\,\pm\,3.35$ \\
 & $r_{z}$ & $-1.12$ & ${}\,\mathbf{\pm\,0.76}$ & $-3.75$ & ${}\,\pm\,1.03$ & $-5.32$ & ${}\,\pm\,1.87$ \\
 & $\tau$ & $204.90$ & ${}\,\mathbf{\pm\,2.72}$ & $203.18$ & ${}\,\pm\,2.91$ & $200.09$ & ${}\,\pm\,3.77$ \\
\bottomrule
\end{tabular*}
\vspace{2pt}
\par\begin{minipage}{\linewidth}\scriptsize\raggedright \textit{Note:} Units and notation as in Table~\ref{tab:modality_radar_main}. -- is unavailable; the Sequence~3 CAD reference is $\bm{p}=\bm{0}$ and $\bm{r}=\bm{0}$.\end{minipage}
\end{table}

After compensating the correlation peak, the event activity follows the reference angular motion. The local time STD ranges from 0.43 to 2.72 ms across recordings; the absolute offsets vary by several milliseconds and are descriptive rather than an independent timing validation. In Seq.~3 the recovered translation is close to the zero CAD reference, whereas Seq.~1 gives a 103.19 cm \(z\) component despite a 4.48 cm local STD. This contrast follows from the measurement model: normal flow is a scalar projection of image-plane motion, and translation is constrained through its coupling with rotation and the lever arm. Strong local conditioning on rotation and time therefore does not imply translation identifiability when the flow directions and angular excitation are insufficient. Rotation estimates agree with the baselines in Seq.~3 and Seq.~4, whereas Seq.~2 departs from both baselines in \(r_y\), consistent with the weak translation excitation of that rotation-dominant recording. The Seq.~3 residual histogram is approximately zero-mean. These are consistency checks, not proof of unbiased extrinsics.

\subsubsection{IMU Calibration}

\begin{figure}[t]
    \centering
    \includegraphics[width=\linewidth]{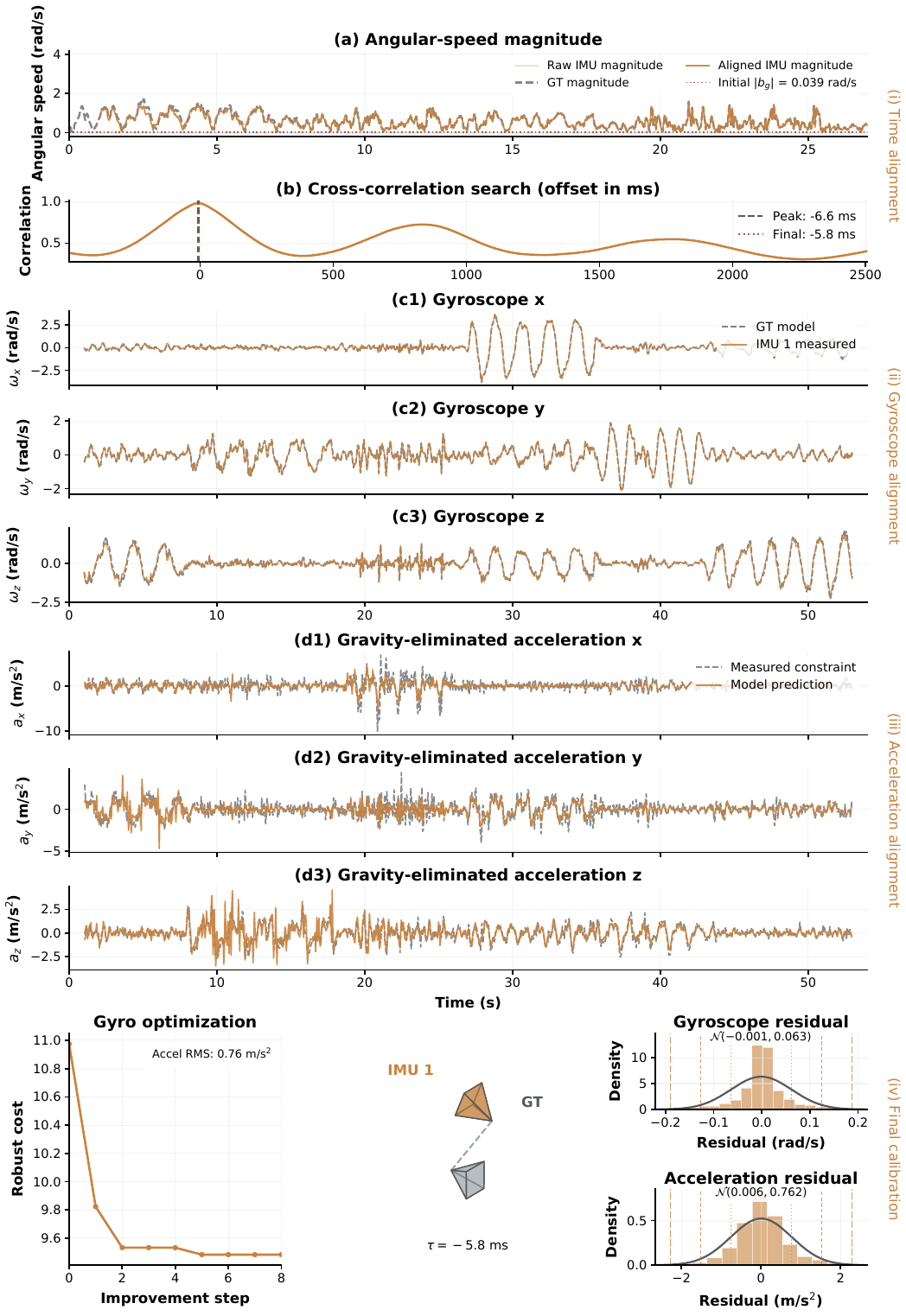}
    \caption{Livox IMU calibration diagnostics for Sequence~3. (a) Angular-speed alignment; (b) offset correlation; (c1)--(c3) gyroscope alignment; (d1)--(d3) acceleration constraints. The bottom row shows optimization, sensor geometry, and residual distributions. The offset is relative to the reference twist frame.}
    \label{fig:imu1_pipeline}
\end{figure}

\begin{figure}[t]
    \centering
    \includegraphics[width=\linewidth]{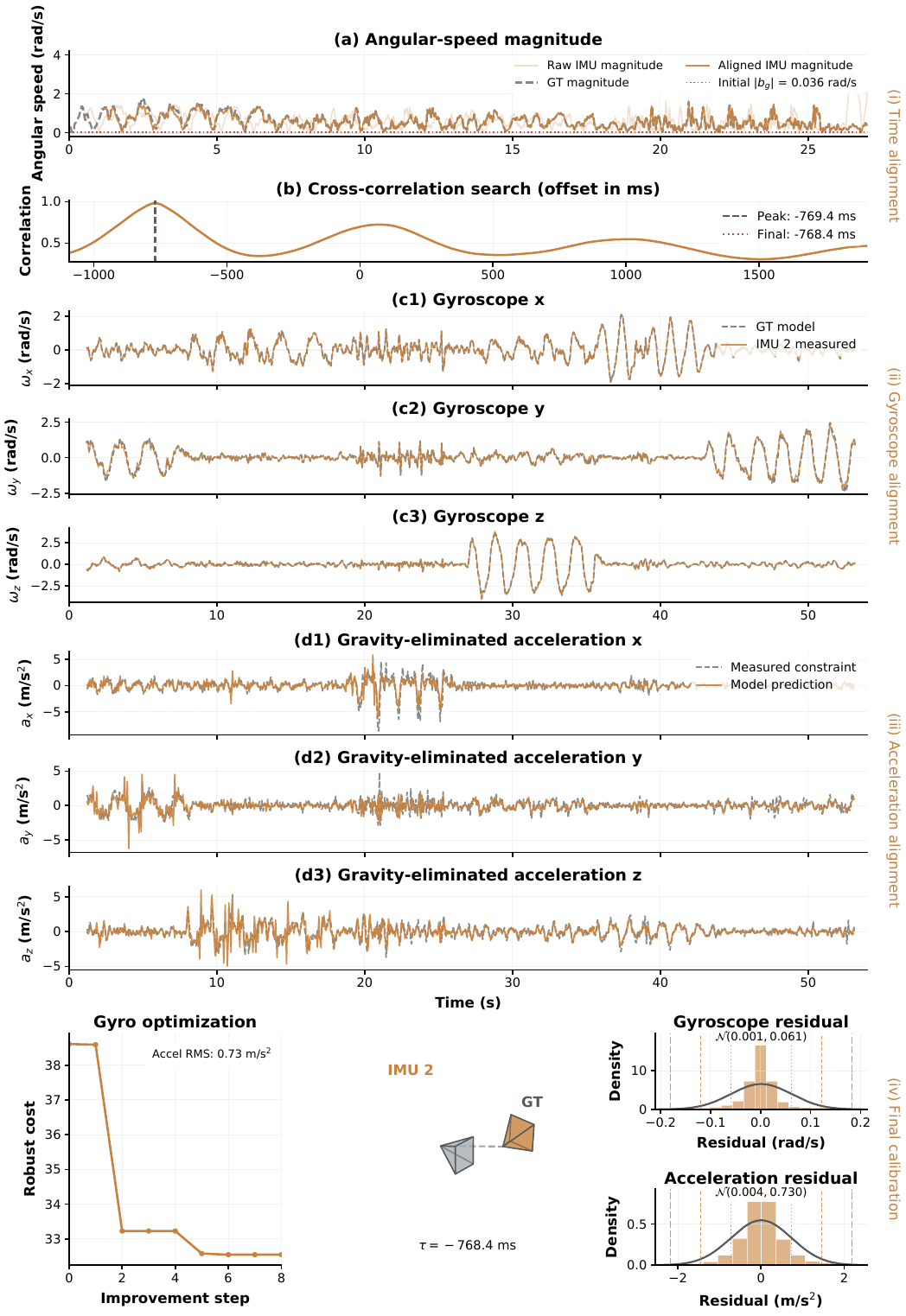}
    \caption{\(\mathcal I_2\) (DVS-IMU) calibration diagnostics for Sequence~3. (a) Angular-speed alignment; (b) offset correlation; (c1)--(c3) gyroscope alignment; (d1)--(d3) acceleration constraints. The bottom row shows optimization, sensor geometry, and residual distributions. The offset is relative to the reference twist frame.}
    \label{fig:imu2_pipeline}
\end{figure}

\begin{figure}[h]
    \centering
    \includegraphics[width=\linewidth]{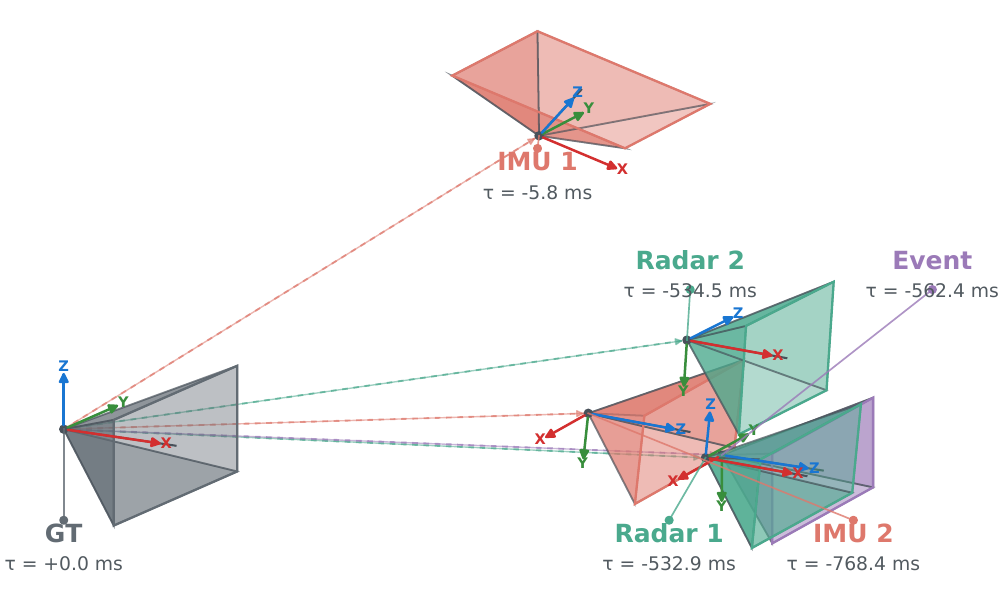}
    \caption{Estimated sensor frames and time offsets for Sequence~3 relative to the reference twist frame. Table offsets are instead relative to \(\mathcal I_2\).}
    \label{fig:seq3_all_sensors}
\end{figure}

Table~\ref{tab:modality_livox_imu} reports the calibration between the Livox IMU \(\mathcal I_1\) and the DVS-IMU reporting frame \(\mathcal I_2\), with Sequence~3 diagnostics in Figs.~\ref{fig:imu1_pipeline} and~\ref{fig:imu2_pipeline} and the estimated layout in Fig.~\ref{fig:seq3_all_sensors}. Our method gives the lowest local uncertainty for most states across the four recordings.

\begin{table}[!t]
\centering
\caption{Inertial calibration results between \(\mathcal I_1\) and \(\mathcal I_2\).}
\label{tab:modality_livox_imu}
\scriptsize
\setlength{\tabcolsep}{1.5pt}
\renewcommand{\arraystretch}{0.92}
\begin{tabular*}{\columnwidth}{@{\extracolsep{\fill}}llr@{}lr@{}lr@{}l}
\toprule
Seq. & State & \multicolumn{2}{c}{Ours} & \multicolumn{2}{c}{MI-Calib~\cite{chen2024mi}} & \multicolumn{2}{c}{iKalibr~\cite{chen2025ikalibr}} \\
\midrule
\multirow{7}{*}{1} & $p_{x}$ & $-0.58$ & ${}\,\pm\,7.87$ & $-5.76$ & ${}\,\pm\,7.28$ & $-3.82$ & ${}\,\pm\,\mathbf{3.63}$ \\
 & $p_{y}$ & $-5.47$ & ${}\,\pm\,7.38$ & $-7.38$ & ${}\,\pm\,\mathbf{1.64}$ & $-9.23$ & ${}\,\pm\,4.25$ \\
 & $p_{z}$ & $-1.96$ & ${}\,\pm\,\mathbf{6.52}$ & $-2.68$ & ${}\,\pm\,6.70$ & $-4.39$ & ${}\,\pm\,7.97$ \\
 & $r_{x}$ & $54.82$ & ${}\,\pm\,\mathbf{0.37}$ & $58.92$ & ${}\,\pm\,0.69$ & $58.29$ & ${}\,\pm\,0.61$ \\
 & $r_{y}$ & $-54.29$ & ${}\,\pm\,\mathbf{0.39}$ & $-57.57$ & ${}\,\pm\,0.93$ & $-56.11$ & ${}\,\pm\,1.32$ \\
 & $r_{z}$ & $76.42$ & ${}\,\pm\,\mathbf{0.33}$ & $70.64$ & ${}\,\pm\,1.08$ & $75.12$ & ${}\,\pm\,0.58$ \\
 & $\tau$ & $815.56$ & ${}\,\pm\,\mathbf{0.36}$ & $816.44$ & ${}\,\pm\,0.68$ & $802.96$ & ${}\,\pm\,4.68$ \\
\midrule
\multirow{7}{*}{2} & $p_{x}$ & $-1.04$ & ${}\,\pm\,1.95$ & $-0.38$ & ${}\,\pm\,\mathbf{1.53}$ & $-0.03$ & ${}\,\pm\,1.77$ \\
 & $p_{y}$ & $-7.15$ & ${}\,\pm\,\mathbf{2.06}$ & $-10.14$ & ${}\,\pm\,3.21$ & $-4.10$ & ${}\,\pm\,4.18$ \\
 & $p_{z}$ & $-3.52$ & ${}\,\pm\,\mathbf{2.90}$ & $-3.95$ & ${}\,\pm\,4.96$ & $-4.94$ & ${}\,\pm\,3.30$ \\
 & $r_{x}$ & $53.78$ & ${}\,\pm\,\mathbf{0.09}$ & $57.37$ & ${}\,\pm\,0.74$ & $59.07$ & ${}\,\pm\,0.55$ \\
 & $r_{y}$ & $-53.03$ & ${}\,\pm\,\mathbf{0.08}$ & $-57.09$ & ${}\,\pm\,0.44$ & $-52.16$ & ${}\,\pm\,0.49$ \\
 & $r_{z}$ & $76.65$ & ${}\,\pm\,\mathbf{0.09}$ & $65.49$ & ${}\,\pm\,0.80$ & $67.05$ & ${}\,\pm\,0.52$ \\
 & $\tau$ & $765.19$ & ${}\,\pm\,\mathbf{0.27}$ & $749.99$ & ${}\,\pm\,0.83$ & $758.89$ & ${}\,\pm\,0.81$ \\
\midrule
\multirow{7}{*}{3} & $p_{x}$ & $-0.70$ & ${}\,\mathbf{\pm\,2.47}$ & $-1.24$ & ${}\,\pm\,3.11$ & $-2.64$ & ${}\,\pm\,3.06$ \\
 & $p_{y}$ & $-7.94$ & ${}\,\mathbf{\pm\,2.29}$ & $-7.39$ & ${}\,\pm\,6.10$ & $-7.67$ & ${}\,\pm\,5.34$ \\
 & $p_{z}$ & $-1.70$ & ${}\,\mathbf{\pm\,3.50}$ & $-6.04$ & ${}\,\pm\,6.83$ & $-5.52$ & ${}\,\pm\,4.74$ \\
 & $r_{x}$ & $53.89$ & ${}\,\pm\,\mathbf{0.07}$ & $56.73$ & ${}\,\pm\,0.11$ & $55.38$ & ${}\,\pm\,0.10$ \\
 & $r_{y}$ & $-52.98$ & ${}\,\pm\,\mathbf{0.06}$ & $-53.91$ & ${}\,\pm\,0.34$ & $-57.33$ & ${}\,\pm\,0.68$ \\
 & $r_{z}$ & $76.64$ & ${}\,\pm\,\mathbf{0.07}$ & $78.48$ & ${}\,\pm\,0.11$ & $79.13$ & ${}\,\pm\,0.45$ \\
 & $\tau$ & $762.56$ & ${}\,\pm\,\mathbf{0.18}$ & $761.44$ & ${}\,\pm\,0.21$ & $764.59$ & ${}\,\pm\,0.28$ \\
\midrule
\multirow{7}{*}{4} & $p_{x}$ & $-1.73$ & ${}\,\pm\,\mathbf{3.10}$ & $2.96$ & ${}\,\pm\,4.27$ & $2.45$ & ${}\,\pm\,6.20$ \\
 & $p_{y}$ & $-8.28$ & ${}\,\pm\,\mathbf{3.34}$ & $-9.67$ & ${}\,\pm\,5.88$ & $-6.83$ & ${}\,\pm\,4.16$ \\
 & $p_{z}$ & $-2.23$ & ${}\,\pm\,\mathbf{4.05}$ & $-6.99$ & ${}\,\pm\,5.69$ & $-5.12$ & ${}\,\pm\,4.92$ \\
 & $r_{x}$ & $53.87$ & ${}\,\pm\,\mathbf{0.07}$ & $54.47$ & ${}\,\pm\,0.11$ & $55.23$ & ${}\,\pm\,0.38$ \\
 & $r_{y}$ & $-53.06$ & ${}\,\pm\,\mathbf{0.06}$ & $-54.32$ & ${}\,\pm\,0.15$ & $-55.52$ & ${}\,\pm\,0.27$ \\
 & $r_{z}$ & $77.59$ & ${}\,\pm\,\mathbf{0.07}$ & $72.35$ & ${}\,\pm\,0.32$ & $74.85$ & ${}\,\pm\,0.49$ \\
 & $\tau$ & $-536.94$ & ${}\,\pm\,\mathbf{0.15}$ & $-533.54$ & ${}\,\pm\,0.28$ & $-536.59$ & ${}\,\pm\,0.33$ \\
\bottomrule
\end{tabular*}
\vspace{2pt}
\par\begin{minipage}{\linewidth}\scriptsize\raggedright \textit{Note:} Units and notation as in Table~\ref{tab:modality_radar_main}. CAD (Seq.~3): $\bm{p}=(-0.34,-6.20,-3.72)$ cm and $\bm{r}=(50.65,-50.65,79.28)^\circ$.\end{minipage}
\end{table}

The gyroscope alignment yields local time-offset STDs of 0.15--0.36 ms across recordings and a rotation STD of about 0.07$^\circ$ in Seq.~3. Rotation and timing uncertainties are generally lower than those of the baselines in Seq.~2 and Seq.~3. The accelerometer branch then uses sliding-window constraints without a global gravity state; its residual couples the specific-force increment to rotation, lever arm, bias, and scale without integrating attitude or gravity over the full sequence. Its \(p_x\) STD decreases from 7.87 cm in Seq.~1 to 2.47 and 3.10 cm in Seq.~3 and Seq.~4, consistent with the excitation change. The accelerometer scales in Table~\ref{tab:ablation_gravity} depart substantially from unity, which explains why removing them degrades the fit. The Seq.~3 residual histograms are approximately zero-mean, supporting model consistency within the tested window model. These are conditional covariance estimates, not errors against independent ground truth.

\subsubsection{Sensor Residuals}
Table~\ref{tab:rmse_comparison_mod} reports RMSE after applying the frozen calibration parameters without refitting. Radar and gyroscope rows use the corresponding velocity and angular-rate constraints. Because the event baselines estimate rotation and time but not translation, the event row compares only angular-velocity alignment with a common front-end estimate; these residuals have different units and do not measure downstream trajectory accuracy.

Both radar residuals and the gyroscope residuals are lowest for our method across the four recordings. In Seq.~3, for example, the radar \(\mathcal R_1\) velocity RMSE is 0.151 versus 0.171 and 0.170 m/s for the baselines, and the gyroscope RMSE is 0.063 versus 0.071 and 0.087 rad/s. The radar advantage is largest in Seq.~1, where the baselines also exhibit the widest extrinsic scatter; once the motion excites the lever arm, the methods explain the shared observations at more similar residual levels. The event angular cue is close, with EvMultiCalib slightly lower in Seq.~3 and Seq.~4, while the acceleration-preintegral comparison is mixed. The event cue may use radar and motion priors, so it is a rotation-and-time consistency check rather than an independent normal-flow evaluation. Overall, these results characterize model consistency; they do not by themselves establish extrinsic accuracy or velocity-estimation gains.

\begin{table}[!t]
\centering
\caption{Residual RMSE on shared observations. Lower is better within each modality; Event angular-velocity RMSE evaluates rotation and time only. CAD* uses Ours' timing and is not ranked.}
\label{tab:rmse_comparison_mod}
\scriptsize
\setlength{\tabcolsep}{1.5pt}
\renewcommand{\arraystretch}{0.92}
\begin{tabular*}{\columnwidth}{@{\extracolsep{\fill}}clrrrr}
\toprule
Seq. & Modality & Ours & Baseline 1 & Baseline 2 & CAD* \\
\midrule
\multirow{5}{*}{1} 
& \(\mathcal{R}_1\) & 0.149 & 0.236 & 0.268 & 0.154 \\
& \(\mathcal{R}_2\) & 0.127 & 0.144 & 0.147 & 0.133 \\
& \(\mathcal{E}_\omega\) & 0.7990 & 0.8165 & 0.8178 & 0.8164 \\
& \(\mathcal{I}_a\) & 1.037 & 1.041 & 1.437 & 1.033 \\
& \(\mathcal{I}_\omega\) & 0.074 & 0.081 & 0.100 & 0.079 \\
\midrule
\multirow{5}{*}{2} 
& \(\mathcal{R}_1\) & 0.097 & 0.170 & 0.108 & 0.109 \\
& \(\mathcal{R}_2\) & 0.107 & 0.133 & 0.135 & 0.109 \\
& \(\mathcal{E}_\omega\) & 1.3902 & 1.6837 & 1.6848 & 1.6841 \\
& \(\mathcal{I}_a\) & 0.627 & 0.614 & 0.645 & 0.624 \\
& \(\mathcal{I}_\omega\) & 0.057 & 0.164 & 0.139 & 0.076 \\
\midrule
\multirow{5}{*}{3} 
& \(\mathcal{R}_1\) & 0.151 & 0.171 & 0.170 & 0.152 \\
& \(\mathcal{R}_2\) & 0.121 & 0.143 & 0.145 & 0.122 \\
& \(\mathcal{E}_\omega\) & 1.59972 & 1.59972 & 1.59848 & 1.60172 \\
& \(\mathcal{I}_a\) & 0.923 & 0.910 & 0.906 & 0.917 \\
& \(\mathcal{I}_\omega\) & 0.063 & 0.071 & 0.087 & 0.078 \\
\midrule
\multirow{5}{*}{4} 
& \(\mathcal{R}_1\) & 0.114 & 0.171 & 0.126 & 0.115 \\
& \(\mathcal{R}_2\) & 0.100 & 0.117 & 0.124 & 0.105 \\
& \(\mathcal{E}_\omega\) & 1.88855 & 1.88766 & 1.88673 & 1.88389 \\
& \(\mathcal{I}_a\) & 1.500 & 1.293 & 1.451 & 1.485 \\
& \(\mathcal{I}_\omega\) & 0.064 & 0.079 & 0.072 & 0.078 \\
\bottomrule
\end{tabular*}
\vspace{2pt}
\begin{flushleft}
\scriptsize
\textit{Note:} Units: radar velocity m/s; event angular velocity rad/s; acceleration preintegral m/s$^2$; gyroscope rad/s. For multiple units, the same modality symbols carry an additional sensor index, e.g., \(\mathcal{I}_{a,1}\) and \(\mathcal{I}_{\omega,2}\); the unindexed \(\mathcal{I}_a\) and \(\mathcal{I}_\omega\) denote the inertial-modality entries of this common-cue comparison. Baseline 1/2 are RIs-Calib~\cite{riscalib2024}/iKalibr~\cite{chen2025ikalibr} (radar), RTEI-Calib~\cite{rtei2024}/EvMultiCalib~\cite{mai2025eventcalib} (event), and MI-Calib~\cite{chen2024mi}/iKalibr~\cite{chen2025ikalibr} (IMU). The event row uses a common angular-rate cue with each method's rotation and time; CAD* uses Ours' timing.
\end{flushleft}
\end{table}

\subsubsection{Relation to Unified Calibration}
\label{subsec:unified_comparison}

iKalibr~\cite{chen2025ikalibr} follows a complementary route: it jointly estimates the body trajectory and all sensor parameters in a continuous-time factor graph and demonstrates one-shot calibration across IMU-only, radar-inertial, LiDAR-inertial, camera-inertial, and combined suites. Its IMU-only and radar-inertial experiments are closest to the present data types, but answer a different question: whether one estimator can recover a complete multi-sensor configuration without an external motion source. Our experiments instead fix a trusted reference twist and ask whether the resulting normal equations expose separate sensor-specific blocks.

This distinction is structural rather than a claim of universal numerical superiority. Joint estimation remains applicable when no trusted reference is available, and it is preferable when the available reference cannot be synchronized or its error model is unknown. Reference conditioning is attractive when a synchronized, independently validated reference exists and modular initialization, fault isolation, or per-sensor recalibration is required. Because the baseline front ends, state sets, objectives, and uncertainty models differ, Tables~\ref{tab:modality_radar_main}--\ref{tab:modality_livox_imu} are not a controlled Pathway~A/Pathway~B experiment; they test external plausibility and local covariance. The clearest empirical signature of the structural argument is that one modality can lose translation observability while another remains well conditioned under the same reference, without a shared latent twist feeding that error back into every sensor block.

\subsection{Ablation Studies}

\subsubsection{Initialization and Refinement}

Table~\ref{tab:optimization_ablation} compares the complete pipeline with initialization-stage ablations and a staged-only variant on Sequence~3. Each ablation skips the specified initializer before the subsequent stages; it is not a reset of a saved final state. Each variant was run once.

\begin{table*}[t]
\centering
\caption{Initialization and refinement ablation on Sequence~3.}
\label{tab:optimization_ablation}
\setlength{\tabcolsep}{4pt}
\renewcommand{\arraystretch}{1.12}
\resizebox{\textwidth}{!}{%
\begin{tabular}{lccccccccr}
\toprule
Configuration & \(\mathcal{R}_1\) & \(\mathcal{R}_2\) & \(\mathcal{E}_\omega\) & \(\mathcal{I}_{\omega,1}\) & \(\mathcal{I}_{a,1}\) & \(\mathcal{I}_{\omega,2}\) & \(\mathcal{I}_{a,2}\) & Refinement status & Time (s) \\
\midrule
Full method & 0.1456 $\rightarrow$ 0.1456 & 0.1126 $\rightarrow$ 0.1126 & 0.1655 $\rightarrow$ 0.1650 & 0.0601 $\rightarrow$ 0.0599 & 0.2496 $\rightarrow$ 0.2498 & 0.0577 $\rightarrow$ 0.0574 & 0.1987 $\rightarrow$ 0.1990 & Converged & 35.1 \\
No time init. & 0.3919 $\rightarrow$ 0.3919 & 0.3581 $\rightarrow$ 0.3581 & 0.1868 $\rightarrow$ 0.1862 & 0.6186 $\rightarrow$ 0.6185 & 0.4554 $\rightarrow$ 0.4553 & 0.6210 $\rightarrow$ 0.6210 & 0.4120 $\rightarrow$ 0.4119 & Converged & 48.6 \\
No rotation init. & 0.1456 $\rightarrow$ 0.1456 & 0.1126 $\rightarrow$ 0.1126 & 0.1964 $\rightarrow$ 0.1966 & 0.0601 $\rightarrow$ 0.0599 & 0.2496 $\rightarrow$ 0.2498 & 0.0577 $\rightarrow$ 0.0574 & 0.1987 $\rightarrow$ 0.1990 & Converged & 43.6 \\
No translation init. & 0.1456 $\rightarrow$ 0.1456 & 0.1126 $\rightarrow$ 0.1126 & 0.1658 $\rightarrow$ 0.1650 & 0.0601 $\rightarrow$ 0.0599 & 0.2496 $\rightarrow$ 0.2498 & 0.0577 $\rightarrow$ 0.0574 & 0.1987 $\rightarrow$ 0.1990 & Converged & 38.9 \\
Zero initialization & 0.5991 $\rightarrow$ 0.5991 & 0.3581 $\rightarrow$ 0.3581 & 0.2037 $\rightarrow$ 0.2026 & 0.6637 $\rightarrow$ 0.6637 & 0.4730 $\rightarrow$ 0.4732 & 0.6210 $\rightarrow$ 0.6210 & 0.4120 $\rightarrow$ 0.4119 & Converged & 52.1 \\
Staged only & 0.1456 & 0.1126 & 0.1655 & 0.0601 & 0.2496 & 0.0577 & 0.1987 & Skipped & 19.0 \\
\bottomrule
\end{tabular}%
}
\par\vspace{2pt}\begin{minipage}{\linewidth}\scriptsize\raggedright \textit{Note:} Entries are residual RMS before and after refinement; staged only reports the staged result. Radar is in m/s, gyroscope in rad/s, event in residual units, and accelerometer in weighted window-constraint m/s$^2$. Both IMUs use 0.10~s windows with a 0.05~s step. Time includes enabled stages; each row is one run, and convergence is a solver status.\end{minipage}
\end{table*}

Skipping time initialization increases the radar and IMU residuals substantially, for example from 0.146 to 0.392 for radar \(\mathcal R_1\) and from 0.060 to 0.619 for the Livox gyroscope \(\mathcal I_{\omega,1}\), even though the solver reports convergence. Skipping rotation initialization raises the event residual from 0.166 to 0.196 and converges to an event rotation 176.2$^\circ$ and a time offset 288.46 ms away from the full-method estimate, indicating a different solution basin. Skipping translation initialization has little effect in this recording. The staged-only result matches the full method at the displayed precision; refinement changes the event state by about 1$^\circ$ and 1 cm and increases runtime from 19.0 to 35.1 s. These are single-run observations, not repeatability or general runtime statistics.

\subsubsection{Bias and Scale}

Table~\ref{tab:sensor_parameter_ablation} compares fixed and estimated bias and scale on Sequence~3, with the extrinsics re-optimized in each configuration.

\begin{table}[t]
\centering
\caption{Bias and scale ablation.}
\label{tab:sensor_parameter_ablation}
\scriptsize
\setlength{\tabcolsep}{2pt}
\renewcommand{\arraystretch}{1.0}
\begin{tabular*}{\linewidth}{@{\extracolsep{\fill}}lccrr@{}}
\toprule
Modality & Bias & Scale & RMSE & $\Delta$ RMSE (\%) \\
\midrule
\multirow{4}{*}{\(\mathcal{R}_1\)} 
& & & 0.15065 & +3.38 \\
& & $\checkmark$ & 0.14694 & +0.83 \\
& $\checkmark$ & & 0.14939 & +2.52 \\
& $\checkmark$ & $\checkmark$ & 0.14572 & +0.00 \\
\midrule
\multirow{4}{*}{\(\mathcal{R}_2\)} 
& & & 0.12069 & +2.87 \\
& & $\checkmark$ & 0.11908 & +1.50 \\
& $\checkmark$ & & 0.11893 & +1.37 \\
& $\checkmark$ & $\checkmark$ & 0.11732 & +0.00 \\
\midrule
\multirow{2}{*}{\(\mathcal{I}_{\omega,1}\)} 
& & -- & 0.06326 & +0.37 \\
& $\checkmark$ & -- & 0.06302 & +0.00 \\
\midrule
\multirow{4}{*}{\(\mathcal{I}_{a,1}\)} 
& & & 0.90796 & +19.10 \\
& $\checkmark$ & & 0.90793 & +19.10 \\
& & $\checkmark$ & 0.76249 & +0.02 \\
& $\checkmark$ & $\checkmark$ & 0.76232 & +0.00 \\
\midrule
\multirow{2}{*}{\(\mathcal{I}_{\omega,2}\)} 
& & -- & 0.06070 & +0.03 \\
& $\checkmark$ & -- & 0.06069 & +0.00 \\
\midrule
\multirow{4}{*}{\(\mathcal{I}_{a,2}\)} 
& & & 0.90962 & +24.59 \\
& $\checkmark$ & & 0.90958 & +24.58 \\
& & $\checkmark$ & 0.73025 & +0.02 \\
& $\checkmark$ & $\checkmark$ & 0.73009 & +0.00 \\
\bottomrule
\end{tabular*}
\par\vspace{2pt}\begin{minipage}{\linewidth}\scriptsize\raggedright \textit{Note:} Checkmark: estimated; blank: fixed. Radar parameters are $\bm{b}_{\mathcal R}^{\mathcal B}$ and $s_{\mathcal R}$; IMU parameters are $\bm{b}_g$, $\bm{b}_a^{\mathcal B}$, and $s_a$. Gyroscope scale is not estimated (--). Units: radar m/s; gyroscope rad/s; accelerometer m/s$^2$. Extrinsics are re-optimized; $\Delta$ RMSE is relative to full estimation.\end{minipage}
\end{table}

For both radars, estimating both parameters gives the lowest residual; scale dominates for \(\mathcal R_1\), whereas bias and scale have comparable effects for \(\mathcal R_2\). For the IMUs, fixing gyroscope bias changes little, whereas fixing accelerometer scale increases the residual by about 19\% for Livox and 25\% for DVS. The estimated accelerometer scales in Table~\ref{tab:ablation_gravity} are far from unity, so treating them as calibration states is important. These are fitting-residual comparisons, not independent validation of the estimated parameters.

\subsubsection{Gravity Modeling}

To examine the gravity treatment, we compare the body-decoupled model with a world-integrated model that explicitly estimates a world-frame gravity vector.

\begin{table}[!t]
\centering
\caption{Gravity model ablation.}
\label{tab:ablation_gravity}
\scriptsize
\setlength{\tabcolsep}{2pt}
\renewcommand{\arraystretch}{1.0}
\begin{tabular*}{\linewidth}{@{\extracolsep{\fill}}llccccc@{}}
\toprule
Modality & Model & States & RMSE & Cond. & Coupling & Scale \\
\midrule
\multirow{2}{*}{\(\mathcal{I}_{a,1}\)} & Body-decoupled (Ours) & 7 & \(\mathbf{0.762}\) & \(\mathbf{3.10}\) & \(\mathbf{0.722}\) & 0.569 \\
& World-integrated & 10 & 0.806 & 13.83 & 0.790 & 0.763 \\
\midrule
\multirow{2}{*}{\(\mathcal{I}_{a,2}\)} & Body-decoupled (Ours) & 7 & \(\mathbf{0.730}\) & \(\mathbf{3.13}\) & \(\mathbf{0.724}\) & 0.643 \\
& World-integrated & 10 & 0.791 & 13.61 & 0.788 & 0.783 \\
\bottomrule
\end{tabular*}
\par\vspace{2pt}\begin{minipage}{\linewidth}\scriptsize\raggedright \textit{Note:} RMSE in m/s$^2$; condition and coupling are numerical diagnostics. The body-decoupled model has seven states without an explicit gravity vector; the world-integrated model adds a three-dimensional gravity vector.\end{minipage}
\end{table}

In Seq.~3, the body-decoupled model uses seven states instead of ten and gives lower acceleration RMSE (0.73--0.76 versus 0.79--0.81 m/s$^2$), condition numbers near 3 instead of 14, and lower coupling. This supports a state-dimension and conditioning advantage in the tested configuration, but a lower residual does not prove the world-integrated model inferior or the extra states redundant; the diagnostics also depend on Jacobian scaling.

\subsubsection{Reference-Input Sensitivity and Motion Degeneracy}

This subsection tests the operating conditions in Sections~\ref{subsec:schur_conditioning} and~\ref{subsec:operating_conditions} by perturbing the reference twist and comparing motion profiles. Tables~\ref{tab:reference_linear_ablation} and~\ref{tab:reference_angular_ablation} report the perturbed-input fit RMSE and the original-reference re-evaluation (clean) RMSE; one value means that both agree at the displayed precision. The former measures fit to the perturbed input, while the latter evaluates the resulting parameters against the original reference. Neither directly measures extrinsic bias.

\begin{table}[!t]
\centering
\scriptsize
\caption{Linear reference-twist sensitivity.}
\label{tab:reference_linear_ablation}
\setlength{\tabcolsep}{2pt}
\renewcommand{\arraystretch}{1.0}
\begin{tabular*}{\linewidth}{@{\extracolsep{\fill}}llccccc@{}}
\toprule
Factor & Level & \(\mathcal{R}_1\) & \(\mathcal{R}_2\) & \(\mathcal{E}_\omega\) & \(\mathcal{I}_{a,1}\) & \(\mathcal{I}_{a,2}\) \\
\midrule
Quant. & 0.01--100$\times$ & 0.151 & 0.121 & 0.112 & 0.762 & 0.730 \\
\midrule
\multirow{3}{*}{Rate} 
& 0.01$\times$ & \twoval{0.327}{0.168} & \twoval{0.321}{0.135} & \twoval{0.084}{0.118} & \twoval{0.962}{0.774} & \twoval{0.926}{0.724} \\
& 0.1$\times$ & \twoval{0.150}{0.151} & \twoval{0.120}{0.121} & \twoval{0.111}{0.112} & \twoval{0.786}{0.771} & \twoval{0.739}{0.749} \\
& 1--100$\times$ & 0.151 & 0.121 & 0.112 & 0.762 & 0.730 \\
\midrule
\multirow{5}{*}{Noise} 
& 0 & 0.151 & 0.121 & 0.112 & 0.762 & 0.730 \\
& 0.001 & 0.151 & 0.121 & 0.112 & \twoval{0.763}{0.762} & \twoval{0.729}{0.729} \\
& 0.01 & 0.151 & 0.121 & 0.112 & \twoval{0.799}{0.761} & \twoval{0.749}{0.709} \\
& 0.1 & \twoval{0.171}{0.151} & \twoval{0.145}{0.121} & \twoval{0.117}{0.112} & \twoval{2.535}{0.761} & \twoval{2.593}{0.706} \\
& 1 & \twoval{0.844}{0.164} & \twoval{0.829}{0.140} & \twoval{0.411}{0.118} & \twoval{24.301}{1.002} & \twoval{24.821}{0.827} \\
\bottomrule
\end{tabular*}
\par\vspace{2pt}\begin{minipage}{\linewidth}\scriptsize\raggedright \textit{Note:} Two-line cells: upper/lower are perturbed-input fit and original-reference re-evaluation RMSE; one value means agreement. Sensor symbols follow Table~\ref{tab:rmse_comparison_mod}. Units: m/s, normal-flow, and m/s$^2$. Noise in m/s; quantization and rate relative to the input. Merged levels agree at the displayed precision.\end{minipage}
\end{table}

\begin{table}[!t]
\centering
\caption{Angular reference-twist sensitivity.}
\label{tab:reference_angular_ablation}
\scriptsize
\setlength{\tabcolsep}{2pt}
\renewcommand{\arraystretch}{1.0}
\begin{tabular*}{\linewidth}{@{\extracolsep{\fill}}llccccccc@{}}
\toprule
Factor & Level & \(\mathcal{R}_1\) & \(\mathcal{R}_2\) & \(\mathcal{E}_\omega\) & \(\mathcal{I}_{a,1}\) & \(\mathcal{I}_{a,2}\) & \(\mathcal{I}_{\omega,1}\) & \(\mathcal{I}_{\omega,2}\) \\
\midrule
Quant. & 0.01--100$\times$ & 0.151 & 0.121 & 0.112 & 0.762 & 0.730 & 0.063 & 0.061 \\
\midrule
\multirow{3}{*}{Rate} 
& 0.01$\times$ & \twoval{0.166}{0.156} & \twoval{0.148}{0.129} & \twoval{0.141}{0.113} & \twoval{1.051}{0.821} & \twoval{0.992}{0.788} & \twoval{0.716}{0.105} & \twoval{0.720}{0.110} \\
& 0.1$\times$ & 0.151 & \twoval{0.120}{0.121} & \twoval{0.113}{0.112} & \twoval{0.748}{0.769} & \twoval{0.714}{0.736} & \twoval{0.122}{0.063} & \twoval{0.121}{0.061} \\
& 1--100$\times$ & 0.151 & 0.121 & 0.112 & 0.762 & 0.730 & 0.063 & 0.061 \\
\midrule
\multirow{2}{*}{Noise} 
& 0--0.01 & 0.151 & 0.121 & 0.112 & 0.762 & 0.730 & 0.063 & 0.061 \\
& 0.1 & 0.151 & 0.121 & \twoval{0.114}{0.112} & \twoval{0.766}{0.762} & \twoval{0.732}{0.731} & \twoval{0.104}{0.063} & \twoval{0.103}{0.060} \\
\bottomrule
\end{tabular*}
\par\vspace{2pt}\begin{minipage}{\linewidth}\scriptsize\raggedright \textit{Note:} Two-line cells: upper/lower are perturbed-input fit and original-reference re-evaluation RMSE; one value means agreement. Sensor symbols follow Table~\ref{tab:rmse_comparison_mod}. Units: m/s, normal-flow, m/s$^2$, and rad/s. Noise in rad/s; quantization and rate relative to the input. Merged levels agree at the displayed precision.\end{minipage}
\end{table}

Quantization has no visible effect at the displayed precision, and a tenfold downsampling to \(0.1\times\) the original rate remains close to the baseline. At \(0.01\times\), the Livox gyroscope \(\mathcal I_{\omega,1}\) fit RMSE rises from 0.063 to 0.716 rad/s, while clean re-evaluation remains about 0.105 rad/s. With 1 m/s linear-velocity noise, the Livox accelerometer \(\mathcal I_{a,1}\) fit RMSE reaches 24.3, but clean re-evaluation is 1.00 versus the 0.76 baseline. Differentiating a noisy reference therefore inflates the fit residual much more than the frozen-parameter residual. The fit/clean distinction does not establish unbiasedness or immunity to reference drift.

The sequence comparison provides the motion-degeneracy counterpart. The radar \(\mathcal R_1\) \(p_x\) STD is 6.08 cm in translation-dominant Seq.~1 and 1.75 and 1.87 cm in axis-separated Seq.~3 and Seq.~4. This trend is consistent with the lever-arm condition in Section~\ref{subsec:schur_conditioning}; Seq.~2 exercises strong rotation with weak linear excitation, while the separate axes in Seq.~3 and Seq.~4 resolve the single-axis null direction. Because duration, scene, and trajectory also differ, the comparison identifies a conditioning effect rather than an isolated causal relation.

These experiments characterize the tested envelope. The joint-estimation baselines are not controlled Pathway~A implementations, since their front ends, objectives, and uncertainty models differ; some radar baseline runs produce implausibly large lever arms, the event baselines do not estimate translation, and no Livox-IMU baseline is available. All sensitivity runs use one synchronized LIO reference, so cross-source repeatability and projection diversity remain outside the tested envelope. The shared-observation RMSE is a consistency check, not a downstream trajectory evaluation.

Taken together, the evidence addresses four different questions. The baseline tables place the estimates in the same range as established methods and show lower local scatter for most proposed blocks; the CAD comparison provides an assembly-level sanity check; the frozen-parameter residuals test whether the estimated states explain shared observations; and the sensitivity and motion studies probe the operating conditions rather than absolute accuracy. The full pipeline requires 35.1 s and the staged-only variant 19.0 s on Sequence~3, which indicates offline engineering feasibility rather than a real-time guarantee. The experiments therefore support the structural decoupling argument and its stated conditions, but they do not establish metrological accuracy, cross-reference repeatability, or a controlled numerical advantage over a Pathway~A implementation with identical objectives.

\clearpage
\section{Conclusion}
This paper addressed reference-conditioned velocity-domain spatiotemporal calibration for a heterogeneous radar--event--inertial suite. The proposed method uses a trusted reference body twist as an exogenous conditioning signal, retains the lever-arm contribution under rotational excitation, and estimates each sensor's rotation, lever arm, time offset, and sensor-specific bias and scale through modality-specific residuals. The probabilistic and local-information analysis shows that, under local identifiability and an independent, synchronized reference, conditioning replaces the shared latent twist by a known input and removes its Schur-complement bottleneck from the parameter normal equations. At the CRLB level, this yields a covariance ordering subject to a bounded reference-bias condition; it does not enlarge the physical observation set. Indoor and forest recordings examine initialization, timing, motion degeneracy, reference-input sensitivity, and shared-observation residual consistency. The evidence supports the stated operating conditions rather than a universal ranking: calibration quality remains conditional on reference accuracy and synchronization, motion excitation, projection diversity, and sensor-model validity. The method is therefore best viewed as a structure-preserving and modular option when such a reference is available, not as a replacement for joint calibration.

\bibliographystyle{ieeetr}
\bibliography{refs}
\end{document}